\documentclass{article}

\PassOptionsToPackage{numbers,compress}{natbib}

 \usepackage[eandd, final]{neurips_2026}
\usepackage[utf8]{inputenc} % allow utf-8 input
\usepackage[T1]{fontenc}    % use 8-bit T1 fonts
\usepackage{hyperref}       % hyperlinks
\usepackage{url}            % simple URL typesetting
\usepackage{booktabs}       % professional-quality tables
\usepackage{amsfonts}       % blackboard math symbols
\usepackage{nicefrac}       % compact symbols for 1/2, etc.
\usepackage{microtype}      % microtypography
\usepackage{xcolor}         % colors

\usepackage{multirow}
\usepackage{arydshln}
\usepackage{pifont}
\usepackage{colortbl}
\usepackage{listings}
\usepackage{xcolor}
\usepackage{tcolorbox}
\usepackage{fvextra}
\usepackage{xspace}
\usepackage{bm}
\usepackage{amsmath}
\usepackage{graphicx}
\usepackage{subcaption}
\usepackage{paralist}

\tcbuselibrary{listings}
\tcbuselibrary{breakable}

\newcommand{\EyeVLM}{{\sc \bf EyeVLM}\xspace}
\newcommand{\mypartitle}[2][]{\par\noindent{\bf #2}}

\newcommand{\SampleImage}{\ensuremath{I}\xspace}
\newcommand{\SampleIn}{\ensuremath{B}\xspace}
\newcommand{\FPrompt}{\ensuremath{Q}\xspace}
\newcommand{\Prompt}{\ensuremath{Prompt}\xspace}
\newcommand{\Tasksubscript}{\ensuremath{Task}\xspace}
\newcommand{\PromptStrategy}{\ensuremath{PStrat}\xspace}
\newcommand{\VLMOut}{\ensuremath{J}\xspace}
\newcommand{\SampleOut}{\ensuremath{O}\xspace}
\newcommand{\LAH}{LAH\xspace}
\newcommand{\LAEO}{LAEO\xspace}
\newcommand{\SA}{SA\xspace}
\newcommand{\GazeFollowingTask}{GFo\xspace}

\newcommand{\FTspecific}{FTspec\xspace}
\newcommand{\FTfull}{AllD\xspace}

\newcommand{\BaselineAcro}{\textsc{PrBase}\xspace}
\newcommand{\VanillaCoTAcro}{\textsc{PrCoTBase}\xspace}
\newcommand{\StructuredCoTAcro}{\textsc{PrCoTStruct}\xspace}

\newcommand{\CoTInContexAcro}{\textsc{PrInContext-CoT}\xspace}

\newcommand{\bestang}[1]{\textbf{#1}$^\circ$}
\newcommand{\ang}[1]{$#1^\circ$}

\newcommand{\code}[1]{
\begin{tcolorbox}[
  colback=gray!10,
  colframe=gray!50,
  boxrule=0.5pt,
  arc=2pt,
  boxsep=2pt,        % padding inside the box
  top=1pt, bottom=1pt, % vertical padding inside
  left=2pt, right=2pt, % horizontal padding inside
  before skip=4pt,   % space above the box
  after skip=4pt     % space below the box
]
#1
\end{tcolorbox}
}
\definecolor{nicegreen}{rgb}{0.1, 0.6, 0.2}
\definecolor{redjmo}{rgb}{0.8, 0.1, 0.1}

\definecolor{noteblue}{rgb}{0.10, 0.35, 0.75}

\title{Eyes on VLM: Benchmarking Gaze Following and Social Gaze Prediction in Vision Language Models}

\author{%
  Hengfei Wang\thanks{Equal contribution.} \\
  Idiap Research Institute\\
  \texttt{hwang@idiap.ch} \\
  \And
  Anshul Gupta\footnotemark[1]\hspace{0.4em}\textsuperscript{\S} \\
  Idiap Research Institute \\
  \texttt{anshul.gupta@tri.global} \\
  \And
  Pierre Vuillecard \\
  Idiap Research Institute \\
  École Polytechnique Fédérale de Lausanne \\
  \texttt{pvuillecard@idiap.ch} \\
  \And
  Jean-Marc Odobez \\
  Idiap Research Institute \\
  École Polytechnique Fédérale de Lausanne \\
  \texttt{odobez@idiap.ch} \\
}

\begin{document}

\maketitle

\begingroup
\renewcommand{\thefootnote}{\S}
\footnotetext{
    This work was conducted while the author was at
    Idiap Research Institute.
}
\endgroup

\vspace{-7mm}
\begin{figure}[h]
    \centering
    \includegraphics[width=1\linewidth]{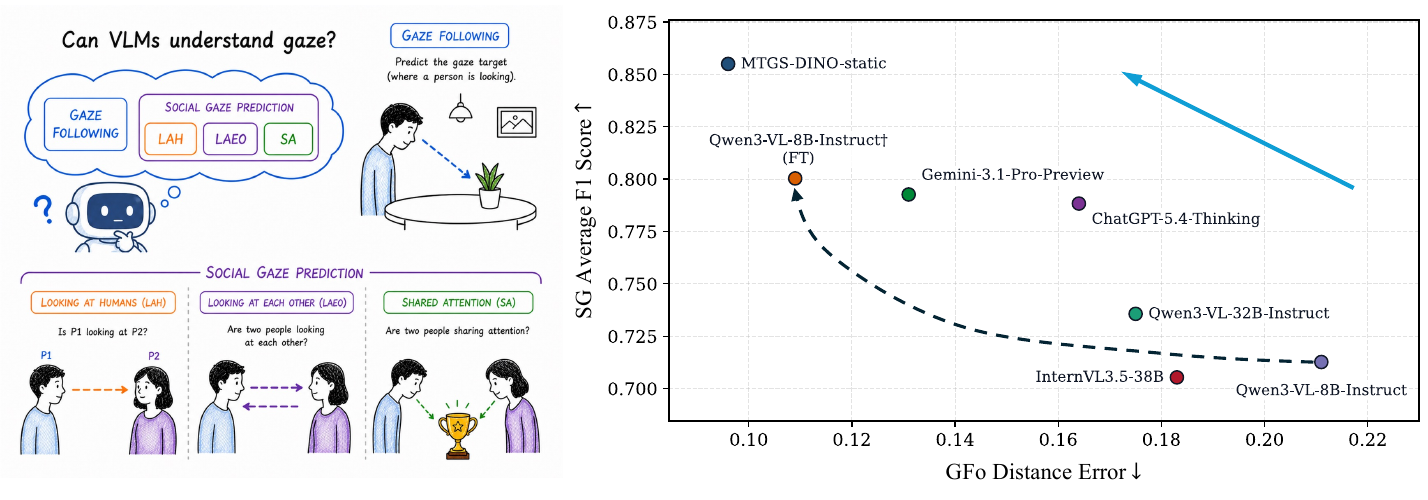}
\caption{\textbf{EyeVLM overview.} Left: Gaze Following (\GazeFollowingTask) and Social Gaze Prediction (SG). Right: model comparison on GazeFollow~\cite{Recasens2015GazeFollow} (\GazeFollowingTask distance error $\downarrow$) and VideoAttentionTarget~\cite{Chong2020VideoAttentionTarget} (SG F1 $\uparrow$).
Results are taken from Table~\ref{tab:vlm_subset_gf_vat}.
Fine-tuning (FT) improves VLMs over dedicated prompting, but they still lag behind specialized pure vision gaze models.
}
    \label{fig:teaser}
\end{figure}

\begin{abstract}
\label{sec:abstract}

Vision-language models (VLMs) have rapidly evolved into general-purpose multimodal reasoners with strong zero-shot generalization. In this context, VLMs could greatly benefit the analysis of human gaze and attention, a central task in human behavior understanding that requires reasoning about the physical scene as well as the activity, interactions, and social context. However, the extent to which VLMs can reliably understand human gaze and related attentional behaviors remains largely unexplored.
In this work, we present \EyeVLM, a unified benchmark for evaluating \emph{gaze understanding} in VLMs along two dimensions: \textit{tasks} and \textit{models}.
To assess gaze understanding capabilities, we focus on two complementary core tasks. The first, \textbf{gaze following}, i.e. predicting the 2D location where a person is looking, has a geometric and visual processing focus, requiring precise understanding of the human face, attention direction, 3D scene structure and spatial grounding of attended targets. 
The second, \textbf{social gaze prediction}, requires social and relational reasoning over multi-person interactions (e.g., mutual gaze and shared attention), and may benefit more from the LLM semantic reasoning capabilities within VLMs.
Regarding models, \EyeVLM evaluates these tasks in two ways: a zero-shot setting with a diverse set of state-of-the-art open- and closed-source VLMs, exploring different prompting strategies; and a fine-tuning approach based on task-specific QA pairs, studying the impact of \textit{model scale} and \textit{data scale}.
As benchmarks, we rely on existing gaze understanding datasets, including a class-balanced social gaze subset that we curate and release,
and compare VLMs with state-of-the-art
purely visual models under the same protocols.
Overall, our results show that current VLMs remain far from specialized models for precise gaze target localization, while they are much closer on social gaze prediction. Fine-tuning substantially reduces the gap with visual models, but improvements are still needed.
Code and benchmark resources are available at
\url{https://github.com/idiap/EyeVLM}.

\end{abstract}

\vspace{-4mm}
\section{Introduction}
\label{sec:introduction}

Understanding gaze is a key component of social cognition, as it provides direct cues about others’ attention and underlying intent \cite{Emery2000,Ryan2025GazeLLE}. It plays a central role in non-verbal communication, supporting the inference of intentions, the regulation of interactions, and the establishment of joint attention \cite{Frischen2007}. Accurate gaze understanding is therefore essential for modeling social behavior and reasoning about interactions and social context.

In computer vision, human gaze understanding has been extensively studied through dedicated tasks such as 3D gaze estimation, gaze following, and social gaze prediction. Gaze following aims to infer a person’s attended target in the scene, typically represented as a 2D point or heatmap. Social gaze prediction focuses on interaction-level phenomena, including mutual gaze (LAEO) \cite{MarinJimenez2021LAEONetPP}, looking-at-person behaviors (LAH) \cite{Gupta2024Unified}, and shared attention (SA) \cite{Fan2018SharedAttention} in multi-person settings. Over the past decade, the field has developed benchmarks and specialized models for these tasks, evolving from early convolutional approaches \cite{Recasens2015GazeFollow,Chong2020VideoAttentionTarget} to more recent transformer-based \cite{Ryan2025GazeLLE} and unified frameworks, including multi-person \cite{Tafasca2024Sharingan} and temporal models \cite{Gupta2024MTGS}, as well as methods incorporating 3D scene representations for improved geometric reasoning.

Despite these advances, it remains unclear to what extent current models achieve true scene understanding. In particular, gaze reasoning requires situated perception, including understanding the person’s activity, modeling interactions with objects or other individuals, and grounding predictions in 3D space. Without such capabilities, generalization and further performance improvements may be limited, and the gap to human-level understanding may persist.

In contrast, vision-language models (VLMs) have recently emerged as general-purpose multimodal reasoners, demonstrating strong zero-shot and few-shot transfer across diverse visual understanding tasks \cite{Radford2021CLIP,Li2023BLIP2,Liu2023LLaVA}. Powered by large-scale pretraining and instruction tuning, modern VLMs can parse complex scenes, capture high-level semantics, and leverage linguistic knowledge about social interactions and human behavior. This suggests that VLMs may be particularly well suited to address the limitations of specialized gaze models, especially for tasks requiring social reasoning and contextual understanding. 

This shift raises a key question: to what extent do VLMs understand human gaze, a fundamental social and perceptual signal underlying attention, communication, and interaction? More broadly, the adoption of VLMs moves the paradigm from task-specific architectures to generalist models guided by prompting and lightweight adaptation, raising the question of whether gaze understanding can be addressed within this framework.
However, despite the maturity of gaze-following and social gaze prediction research, a comprehensive evaluation of modern VLMs on these tasks is still lacking. In particular, we lack a unified study covering (i) diverse gaze tasks and datasets, (ii) a representative set of open and closed-source models under consistent protocols, (iii) the impact of prompting and adaptation strategies, and, importantly, (iv) fair comparisons with state-of-the-art pure visual models.
Existing gaze literature focuses on specialized models, while VLM behavioral studies rarely isolate gaze as a standalone capability, instead treating it implicitly within broader visual reasoning tasks \cite{huang2023kosmos,thumu2026socialcaptionevaluatingsocial}. As a result, it remains unclear what VLMs can achieve out of the box, when and why they fail, and how they should be effectively prompted or adapted for gaze understanding.

To address this gap, we introduce \EyeVLM, a benchmark designed to assess gaze understanding in VLMs (Fig.~\ref{fig:teaser}). It is organized along two main dimensions: tasks and models.

First, at the task level, we study two representative problems: gaze following and social gaze prediction, which exhibit different scene understanding requirements. Gaze following relies on gaze direction perception, 3D scene structure, and geometric reasoning to localize attended targets within a situated context (e.g., activity, gestures). In contrast, the social gaze prediction task emphasizes relational reasoning over multi-person interactions, such as predicting gaze relationships between individuals. EyeVLM standardizes task interfaces so they can be queried through language and enables direct comparison with state-of-the-art visual models.
We deliberately scope the benchmark to these two third-person, frame-based tasks: gaze following is the most studied scene-level gaze task; social gaze adds a relational task without coordinate outputs. Both have established datasets and strong specialized models as reference points.

Second, at the model level, we conduct zero-shot evaluations of a diverse set of state-of-the-art VLMs, including both open and closed-source models. We study multiple prompting strategies (Chain-of-Thought, In-Context learning), including variations in instruction templates and reasoning cues. Given the subtle visual and relational nature of gaze understanding, prompt design is expected to significantly influence both task interpretation and output structure.
Beyond zero-shot evaluation, EyeVLM also probes adaptability. We construct task-specific QA pairs for supervised fine-tuning and adapt representative open VLM families (Qwen \cite{qwen3vltechnicalreport}
) to assess how fine-tuning affects gaze understanding. 
We further analyze the impact of using multiple datasets or tasks during training.

\EyeVLM is designed as a diagnostic tool. By jointly analyzing results across tasks, models, and prompting strategies, we aim to characterize the current capabilities and limitations of VLMs for gaze understanding.
In summary, our main contributions are threefold:
\begin{compactitem}
\item A unified benchmark for gaze understanding in VLMs, covering gaze following and social gaze prediction with standardized language-based interfaces and comparison to vision-based models.

\item Analysis of zero-shot performance and prompting.
We evaluate diverse VLMs and study the impact of prompting strategies (Chain-of-Thought, In-Context) on gaze understanding.

\item Study of fine-tuning. 
We construct QA pairs for adaptation and analyze how fine-tuning, model scale, and data scale influence performance.
\end{compactitem}

\section{Related Work}

\mypartitle{Gaze Following}
Typical approaches to gaze following adopt a two-branch architecture: one branch analyzes the broader scene to identify salient objects or regions, while the other analyzes the head of the person of interest to estimate gaze direction. Information from the two branches is then fused to produce the final prediction \cite{Recasens2015GazeFollow, fang2021dual, Chong2020VideoAttentionTarget, lian2018believe, gupta2022modular, jin2022depth, tafasca2023childplay}. Subsequent work has shown the benefits of incorporating additional modalities, such as scene depth \cite{fang2021dual, gupta2022modular, tafasca2023childplay} and human pose \cite{gupta2022modular}. Other studies further explored temporal modeling \cite{Chong2020VideoAttentionTarget, Miao2023PDP}, but with no or only marginal gains over frame-based counterparts, so most methods remain frame-based. 
Additionally, following Sharingan \cite{Tafasca2024Sharingan}, the first work demonstrating that transformer based architectures can be leveraged for the task, 
multi-person transformer based architectures have enabled single-pass prediction for multiple individuals~\cite{Tafasca2024Sharingan, Gupta2024MTGS}, added object semantic predictions \cite{tafasca_towards_2024}, 
while Gaze-LLE \cite{Ryan2025GazeLLE} leveraged large-scale learned visual encoders with lightweight decoding, highlighting the value of foundation models for gaze understanding, 
as also demonstrated in \cite{mtgs-static-vsgaze}, the DinoV2 version of \cite{Gupta2024MTGS}. 
However, these advances largely assume task-specific training and outputs, motivating EyeVLM to systematically evaluate how general-purpose VLMs handle gaze following under varying models, prompts, and adaptation settings.

\mypartitle{Social Gaze Prediction}
Early works like \cite{funes_semi-automated_2013} rely on precise gaze geometric models applied to dedicated settings for social gaze encoding. 
More recent works in unconstrained videos formulated mutual gaze detection using head pose and geometry cues~\cite{MarinJimenez2013LAEO}.
LAEO-Net and LAEO-Net++ introduced track-based spatiotemporal deep models to improve robustness beyond frame-wise decisions~\cite{MarinJimenez2019LAEONet,MarinJimenez2021LAEONetPP}, and were followed by other deep learning based models \cite{Doosti_Chen_Vemulapalli_Jia_Zhu_Green_2021, cantarini2021hhpnet}.
Fan et al. \cite{Fan2018SharedAttention} introduced a model for in-the-wild shared attention estimation by combining gaze cones from individuals with object region proposals in the scene.
Attention Flow learns shared attention end-to-end, highlighting the benefit of modeling collective focus rather than post-processing individual gaze~\cite{Sumer2020AttentionFlow}.
More recent unified formulations jointly predict gaze targets and social gaze labels~\cite{Gupta2024Unified},
and temporal multi-person frameworks further connect gaze following with social gaze prediction at scale~\cite{Gupta2024MTGS}.
However, existing methods remain specialized and dataset-dependent, motivating EyeVLM to evaluate VLMs on social gaze prediction across model families, prompt strategies, and fine-tuning settings.

\mypartitle{Vision-Language Models}
Vision-language models (VLMs) have shown strong generalization across visual reasoning tasks through large-scale image-text pretraining and instruction tuning~\cite{Radford2021CLIP,Li2023BLIP2,Liu2023LLaVA}.
However, gaze understanding remains a fine-grained and socially grounded capability that is rarely isolated in standard VLM benchmarks.
Recent studies have begun to explore the connection between VLMs and gaze understanding.
Gupta et al.~\cite{gupta2024exploring} use VLMs as zero-shot contextual cue extractors and inject the predicted cues into a specialized gaze following model.
GazeLLM~\cite{yang2025gazellm} employs a pretrained LLM to perform spatial-semantic reasoning over structured object representations, improving specialized vision gaze following architecture through a plug-and-play fusion module.
GazeVLM~\cite{mathew2025gazevlm} develops a dedicated VLM for multi-task gaze understanding with RGB, depth, and textual prompts.
Despite these efforts, existing works mainly use VLMs as auxiliary modules, construct gaze-specific VLMs, or improve specialized gaze models through language reasoning. A recent exception is VL4Gaze~\cite{wang2025vl4gaze}, which formulates gaze following as a VQA-style problem and constructs a multi-task benchmark covering gaze description, localization, and ambiguity recognition.

However, none of these works consider social gaze prediction. Furthermore, none of them compare current general-purpose VLMs with specialized gaze models under both zero-shot and fine-tuning settings.
As a result, it remains unclear where modern VLMs stand on gaze understanding tasks, which tasks they can solve directly, and how much adaptation is needed to close the gap.
This positioning is a necessary foundation for further applying VLMs to gaze understanding.
Motivated by this gap, \EyeVLM provides a unified evaluation of VLMs on gaze following and social gaze prediction across tasks, models, prompts, and adaptation settings.

\section{EyeVLM Framework}

We present \EyeVLM, a unified evaluation framework for investigating and benchmarking gaze understanding in VLMs, including comparisons with state-of-the-art vision-based models.
In Section~\ref{sec:task_formulation}, we introduce the task formulations. Section~\ref{sec:prompt_protocol} describes and motivates the prompting strategies we study, while Section~\ref{sec:finetuning_protocol} details our VLM fine-tuning approach.

\subsection{Task Formulation}
\label{sec:task_formulation}

We formulate both our tasks into a unified VLM-compatible visual question answering setting.
Each sample consists of an image \SampleImage and data \SampleIn that specifies the subject(s) of interest. 
This information is used to generate a prompt \Prompt based on a specific template \FPrompt
that depends on the task and prompting strategy \PromptStrategy, and further 
passed through a VLM to generate a textual output \VLMOut that includes a structured JSON response, enabling automatic parsing to generate the 
result of interest \SampleOut and evaluation.
Formally, we have: 
\begin{align*}
\Prompt & = \FPrompt^{\Tasksubscript,\PromptStrategy}(\SampleImage, \SampleIn)\\ 
\SampleOut & = Parse^{\Tasksubscript}(\VLMOut) \mbox{ with } \VLMOut  = VLM(\Prompt) 
\end{align*}

\mypartitle{Gaze Following.}
In that case, the input \SampleIn simply consists of the 
subject head bounding box $b = [x_{min}, y_{min}, x_{max}, y_{max}]$, while the output is 
$\SampleOut=(p_{\mathrm{io}}, (x, y))$, which comprises 
the probability $p_{\mathrm{io}} \in [0,1]$ 
that the gaze target lies inside the visible image area, 
and $(x,y) \in [0,1]^2$ denotes the normalized gaze point. 
Note that for Qwen models, we use $(x,y) \in [0,1000]^2$ in prompts and outputs, as Qwen models are trained to use this scale for coordinate representation~\cite{qwen3vltechnicalreport}
; predictions are rescaled to $[0,1]^2$ for evaluation.

\mypartitle{Social Gaze.}
Here the input \SampleIn comprises the two head bounding boxes $b_1,b_2$ associated with the social relation that is being tested: 
\LAEO, looking at each other, 
\SA, shared attention between the two persons, 
\LAH, is person $b_1$ looking at person $b_2$'s face.
In all three cases, \SampleOut only comprises
$p_{\mathrm{sg}} \in [0,1]$, which
denotes the confidence that the queried
 social gaze behavior is true.

\subsection{Zero-shot Prompt Protocols}
\label{sec:prompt_protocol}

Many studies have shown that LLMs may perform better with different prompting strategies, especially in domains requiring logical reasoning such as mathematics and code generation \cite{wei2022chain,kojima2022large}. However, their use in VLMs has shown mixed results: while some works demonstrate improved accuracy \cite{lu2022learn,zhang2024multimodal}, others show that they can hurt performance, especially when perception is uncertain \cite{wu2026better}, as it can be in our case.
As our goal is not only to measure zero-shot accuracy, but also to understand which aspects of the gaze tasks are already well captured by VLMs and which are not, we design four prompting strategies with progressively increasing levels of guidance. These strategies cover the two main prompting paradigms (Chain-of-Thought and in-context learning, also known as few-shot prompting), and allow us to systematically analyze prompt sensitivity and impact on performance measures.
Below, we describe their main features, first focusing on gaze following, before detailing how they are adapted to social gaze tasks. Details are in Appendices~\ref{app:prompts_gf} and \ref{app:prompts_sg}.

\mypartitle{\BaselineAcro.} 
This prompt comprises three components: a role description, a task description, and an input–output format description. The role description defines the model identity and provides the general semantic space for task-relevant behaviors:
\code{"You are a vision assistant specializing in human gaze analysis."}
The task description specifies the semantic goal of the queried task. For gaze following, this consists in asking where a person of interest is looking, including (i) the probability that the person's gaze target is within the frame, and (ii) the location of the gaze point within the scene. However, these two outputs are inherently coupled: if the model predicts that the person is looking outside the frame, it may trivially output a default gaze point (e.g., $[0,0]$) rather than reasoning about the most probable gaze location. 
To address this coupling, we design the prompt to ensure that VLMs predict a plausible gaze point even when the person of interest is likely looking outside the frame.
Finally, the format description explicitly constrains the expected JSON output.

\mypartitle{\VanillaCoTAcro.} 
Chain-of-Thought (CoT) is a prompting strategy that encourages VLMs to perform step-by-step reasoning about a given task. While it is widely used to elicit intermediate reasoning, its benefits can extend beyond simple logical decomposition.
In gaze understanding, effective prediction requires integrating multiple heterogeneous cues, including visual perception (e.g., head pose and eye direction), scene semantics (e.g., objects and interactions), and 3D spatial reasoning (e.g., relative positions and visibility). 
Such multi-factor reasoning is often implicit and not directly enforced by standard prompts. CoT provides a mechanism to externalize and structure this process, guiding the model to sequentially attend to relevant cues and intermediate decisions. This is particularly important for VLMs, where reasoning must bridge visual and linguistic modalities.
To evaluate whether VLMs can leverage such structured reasoning, we first consider a vanilla CoT prompt that augments the base prompt with an explicit instruction encouraging step-by-step analysis:
\code{"In order to achieve your task, you can first do a step-by-step analysis."}
This variant allows us to assess whether explicitly eliciting intermediate reasoning improves performance on gaze understanding, and more importantly, whether VLMs can effectively ground this reasoning in visual evidence rather than relying on spurious or language-driven cues.

\mypartitle{\StructuredCoTAcro.}
This prompt provides a more structured reasoning scaffold. 
Its motivation comes from the fact that while naive Chain-of-Thought prompting can be ineffective in VLMs, recent works suggest structured or guided variants significantly improve reasoning, particularly for spatial and visually grounded tasks \cite{chen2024vcot,ge2023cotprompt,ji2025spatialcot} which share similarities with our task. 
We therefore guide the model through a sequence of intermediate questions, including: 
(1) describing the scene, 
(2) identifying the subject's activity, 
(3) inferring gaze direction from head and eye cues, 
(4) enumerating plausible gaze targets inside the field of view, 
(5) describing spatial relations between the subject and candidate targets, 
and (6) synthesizing these cues into the final prediction.
This prompt is intended to encourage models to attend to information that is particularly relevant for gaze reasoning.

\mypartitle{\CoTInContexAcro.}
Studies have shown that directly providing examples to VLMs can be more effective than relying solely on explanation-only prompts, especially when these examples include CoT-guided reasoning \cite{chen2024vcot}. To this end, this prompting strategy relies on two carefully selected in-context examples comprising an image, the question, the CoT process applied to that image, and the corresponding structured output. We manually write the analysis process for each example, considering all aspects mentioned in the \StructuredCoTAcro prompt.
Samples are from the training sets. This variant tests whether exemplars help the model internalize both the expected reasoning pattern and the required output format.

\mypartitle{Social Gaze Prompts.}
They follow the same patterns as the gaze following ones. 
Only the task is adapted and defined as whether the queried pair of people exhibit a social gaze relation: looking-at-each-other (LAEO), shared attention (SA), or looking-at-humans (LAH).
Chain-of-Thought and In-Context prompts are adapted accordingly for each of the social task.

\subsection{Supervised Fine-Tuning}
\label{sec:finetuning_protocol}

While the previous section explores how far VLMs can be pushed through prompting alone, we now investigate whether supervised fine-tuning can further specialize VLMs for gaze understanding. Despite their strong general-purpose multimodal capabilities, VLMs are not explicitly trained to capture gaze and attention cues, which are critical for our tasks. We therefore study whether task-specific supervision can bridge this gap.

\mypartitle{QA Pair Construction.}
Fine-tuning VLMs requires text-based QA pairs, while existing datasets for gaze following and social gaze prediction do not provide such annotations. To address this, we construct QA pairs for both gaze understanding tasks. 
As described in Section~\ref{sec:task_formulation}, each query specifies the task and localizes the person(s) of interest via subject head bounding box(es), combining a task-specific question template with the corresponding subject location.
For gaze following, we further decompose the prediction into two separate queries: an \texttt{inout} query and a \texttt{gaze\_point} query. 
We adopt this decoupled design because jointly predicting both outputs may introduce undesirable causal dependencies in the model responses, as discussed in Section~\ref{sec:prompt_protocol}. Querying these variables separately alleviates this issue and leads to more stable predictions.
Accordingly, the task-specific question templates take the following form:
\begin{tcolorbox}[breakable, colback=gray!10, colframe=gray!50, boxrule=0.5pt, arc=2pt]
\textbf{Gaze Following and Social Gaze Prediction}
\begin{compactitem}
    \item \textit{In/Out Query:} "Is the gaze target of \texttt{<OnePerson>} inside the frame?"
    \item \textit{Gaze Point Query:} "Where is \texttt{<OnePerson>} looking at?"
    \item \textit{LAEO:} "Are the \texttt{<TwoPersons>} in the image looking at each other?"
    \item \textit{LAH:} "Is \texttt{<OnePerson>} looking at \texttt{<OnePerson>}?"
    \item \textit{SA:} "Do \texttt{<TwoPersons>} appear to be sharing attention?"
\end{compactitem}

\textbf{\texttt{<OnePerson>}}: the person whose head is within the bounding box $[x_{\min},\, y_{\min},\, x_{\max},\, y_{\max}]$

\textbf{\texttt{<TwoPersons>}}: the two individuals whose heads are enclosed by the bounding boxes 
    $[x_{\min}^{(1)},\, y_{\min}^{(1)},\, x_{\max}^{(1)},\, y_{\max}^{(1)}]$ 
    and 
    $[x_{\min}^{(2)},\, y_{\min}^{(2)},\, x_{\max}^{(2)},\, y_{\max}^{(2)}]$
\end{tcolorbox}

These are simple examples. To improve generalization, we design diverse question templates for each query type while preserving semantic consistency. In addition, we introduce varied linguistic expressions to specify subject location. During QA pair generation, these components are sampled independently and then combined to form the final queries, significantly increasing their diversity. 
Further details are provided in Appendix~\ref{app:qa}.
For the expected answers, we adopt a structured JSON format, as described in Section~\ref{sec:prompt_protocol}, to ensure consistency and facilitate learning.
Each training sample contains a single task-specific QA pair, without a system prompt. Complete QA examples for each task are given in Appendix~\ref{app:qa_examples}.

\mypartitle{Adaptation Setup.}
We adopt the standard LoRA-based supervised fine-tuning 
(rank 8, batch 64) using our constructed QA pairs, 
optimizing the standard next-token prediction objective. 
We resize images and constrain the total number of pixels to 200{,}704 ($448 \times 448$) to align with other gaze works while preserving the aspect ratio of the images. More details in Appendix \ref{sec:appendixadaptation}.

\section{Experiments}

\subsection{Experimental Protocol}
\label{sec:experimentalprotocol}

We propose an evaluation protocol enabling
us to compare heterogeneous VLMs under a unified interface, despite differences in architecture, training data, and native output behavior.
This includes comparisons with purely vision-based models.

\mypartitle{Datasets.}
We rely on three datasets, \textbf{GazeFollow (GF)}~\cite{Recasens2015GazeFollow}
, \textbf{VideoAttentionTarget (VAT)}~\cite{Chong2020VideoAttentionTarget}
, and \textbf{ChildPlay (CP)}~\cite{tafasca2023childplay}
, for training and evaluation. These datasets cover complementary settings, including static images, dynamic videos, and natural child–adult interactions, enabling a broad assessment of model performance on both gaze following and social gaze prediction tasks. See Appendix~\ref{sec:appendixdataset} for more details.
In their original versions, these datasets provide annotations for gaze following, including head bounding boxes, gaze target locations, and in-frame versus out-of-frame labels (for VAT and ChildPlay). VSGaze \cite{Gupta2024MTGS} extends the annotations of several datasets \cite{Chong2020VideoAttentionTarget, tafasca2023childplay, MarinJimenez2019LAEONet, Fan2018SharedAttention} to include social gaze behaviors. In this work, we use the extended annotations on VAT and ChildPlay to construct the social gaze prediction benchmark.

\mypartitle{Benchmarks.}
For the gaze following task, we follow the standard train/test splits and protocols defined by each dataset, which keeps our results directly comparable with prior work.
For the social gaze prediction task, the original data exhibit substantial class imbalance between positive and negative samples across the three categories, namely \LAEO, \LAH, and \SA, as well as significant redundancy due to dense temporal annotations which can bias training and evaluation. To address these issues, we re-sample the data within the original splits and construct a balanced subset in which positive and negative samples are evenly distributed across all three categories.
Both training and evaluation with VLMs are conducted on this balanced subset, and specialized baselines are re-evaluated on the same test subset. Detailed sampling procedures and the resulting data distributions are provided in the supplementary material (Appendix~\ref{supp:social_gaze_sampling}).

\begin{table*}[t]
\caption{\textbf{Comparison of prompting strategies for gaze following and social gaze prediction.} Results are obtained with \textit{Qwen3-VL-8B-Instruct}~\cite{qwen3vltechnicalreport}
, evaluated on GazeFollow~\cite{Recasens2015GazeFollow} (GF) and VAT~\cite{Chong2020VideoAttentionTarget} (SG). CoT-based prompting significantly improves gaze following, while offering limited or inconsistent gains for social gaze prediction. 
Best in bold, second-best underlined.
}
\label{tab:zero-shot prompts}
\centering
\footnotesize
\setlength{\tabcolsep}{4pt}
\renewcommand{\arraystretch}{1.05}
\begin{tabular}{lcccccc}
\toprule
\multirow{2}{*}{Prompt} & \multicolumn{3}{c}{Gaze Following} & \multicolumn{3}{c}{Social Gaze Prediction} \\
\cmidrule(lr){2-4} \cmidrule(lr){5-7}
& Avg L2 $\downarrow$ & Min L2 $\downarrow$ & Ang Err $\downarrow$ & F1$_{\mathrm{LAH}}\uparrow$ & F1$_{\mathrm{LAEO}}\uparrow$ & F1$_{\mathrm{SA}}\uparrow$\\
\midrule
\BaselineAcro & 0.332 & 0.240 & $48.5^\circ$ & \textbf{0.699} & \textbf{0.794} & \underline{0.630} \\
\VanillaCoTAcro & 0.244 & 0.165 & $36.2^\circ$ & 0.691 & 0.749 & 0.608 \\
\StructuredCoTAcro & \underline{0.224} & \underline{0.153} & \underline{\ang{33.5}} & \underline{0.695} & \underline{0.768} & \textbf{0.633} \\
\CoTInContexAcro & \textbf{0.215} & \textbf{0.143} & \textbf{30.7}$^\circ$ & 0.643 & 0.756 & 0.538 \\
\bottomrule
\end{tabular}
\vspace{-7mm}
\end{table*}

\mypartitle{Metrics.}
For gaze following, we follow the standard evaluation protocol and report metrics commonly used in prior work: the average L2 distance between the predicted gaze point and the ground-truth target (with locations expressed in the $[0,1]\times[0,1]$ normalized image space), and the average precision $AP_{in/out}$ for in/out prediction. On the GazeFollow dataset, we additionally report the minimum distance between the predicted gaze point and multiple ground-truth targets.
We further report the 2D angular error between the predicted and reference gaze directions, which measures how accurately the model infers head pose and gaze direction.
For social gaze prediction, we report precision, recall, and F1 score for the three categories: LAH, LAEO, and SA.

\mypartitle{Output Parsing.}
We parse the JSON output to extract the predicted variables. To keep every test sample, a fallback defined a priori maps malformed outputs to defaults ($0.5$ for \texttt{inout}, $(0.5, 0.5)$ for \texttt{gaze\_point}, $0.5$ for social gaze). In practice it has no effect: parsing succeeded for all zero-shot outputs and failed once across all fine-tuned evaluations (Appendix~\ref{app:parsing}), so excluding failures leaves all metrics unchanged.

\mypartitle{Fine Tuning Approaches.}
To follow recent gaze following work \cite{Gupta2024MTGS,Ryan2025GazeLLE}, we  considered several alternatives. 
First (\textbf{\FTspecific}\footnote{An exception is the MTGS model \cite{Gupta2024MTGS}. We use an updated version \cite{mtgs-static-vsgaze}  that leverages a DINOv2 backbone. 
This model is trained on the GF dataset and fine-tuned on VSGaze. We denote this as \FTspecific$\!^{\ddagger}$.}  approach), fine-tune the VLM on the gaze following task using the GF dataset, and then further fine-tune it on the target task and dataset (e.g., gaze following on VAT).
We also evaluate whether (i) training with more data improves performance, denoted by \textbf{\FTfull} when using all three datasets (GF, VAT, CP) simultaneously; and (ii) joint training on multiple tasks improves results, as shown in prior work \cite{Gupta2024MTGS}. We denote these settings as \textbf{\GazeFollowingTask} when training on the gaze following task only, and \textbf{\GazeFollowingTask+SG} when adding social gaze supervision. 
In the \textbf{\FTfull/\GazeFollowingTask+SG} setting, all QA pairs of all tasks are mixed in one training set (Appendix~\ref{app:qa_examples}).

\begin{table*}[t]
\caption{Gaze following results on the GF and VAT datasets.
We report results in three main blocks.
Top: specialized vision-based gaze following methods ($^{*}$re-evaluated, $^{\#}$reimplementation version from Gaze-LLE).
Middle: zero-shot VLM methods (with \CoTInContexAcro prompt).
Bottom: fine-tuned VLMs, $^{\dagger}$with different training strategies (see Section~\ref{sec:experimentalprotocol}.)}
\label{tab:sota_gf}
\centering
\renewcommand{\arraystretch}{1.12}
\footnotesize
\setlength{\tabcolsep}{4pt}
\renewcommand{\arraystretch}{1.05}
\begin{tabular}{l c c c c c c c c}
\toprule
\multirow{2}{*}{Method} & \multirow{2}{*}{FT Strategy} & \multicolumn{3}{c}{GazeFollow} & \multicolumn{4}{c}{VideoAttentionTarget} \\
\cmidrule(lr){3-5} \cmidrule(lr){6-9}
& & Avg L2$\downarrow$ & Min L2$\downarrow$ & Ang Err$\downarrow$ & L2$\downarrow$ & Ang Err$\downarrow$ & AP$_{\text{in/out}}$$\uparrow$ & F1$_{\text{in/out}}$$\uparrow$ \\
\midrule
\textit{One Human} & -- & 0.096 & 0.040 & $11.0^\circ$ & 0.051 & -- & 0.925 & -- \\
\midrule
Tonini~\cite{tonini2022multimodal} & \FTspecific  & 0.141 & -- & -- & 0.125 & -- & 0.742 & -- \\
Hu~\cite{hu2022gaze} & \FTspecific & 0.128 & 0.069 & -- & 0.118 & -- & 0.881 & -- \\
Gupta~\cite{gupta2022modular} & \FTspecific & 0.114 & 0.056 & -- & 0.110 & -- & 0.879 & -- \\
Horanyi$^{\#}$~\cite{horanyi2023they} & \FTspecific & 0.196 & 0.127 & -- & 0.199 & -- & 0.800 & -- \\
Miao~\cite{Miao2023PDP} & \FTspecific & 0.123 & 0.065 & -- & 0.109 & -- & \textbf{0.908} & -- \\
Tafasca~\cite{tafasca2023childplay} & \FTspecific & 0.122 & 0.062 & -- & 0.109 & -- & 0.834 & -- \\
Tafasca~\cite{Tafasca2024Sharingan} & \FTspecific & 0.113 & 0.057 & -- & 0.107 & -- & 0.891 & -- \\
Gaze-LLE (ViT-B)$^{*}$~\cite{Ryan2025GazeLLE} & \FTspecific & 0.104 & 0.045 & \ang{12.4} & 0.107 & \ang{14.9} & 0.898 & 0.841 \\
Gaze-LLE (ViT-L)$^{*}$~\cite{Ryan2025GazeLLE} & \FTspecific & 0.098 & \textbf{0.041} & \ang{11.6} & 0.103 & \ang{13.3} & 0.903 & \textbf{0.875} \\
MTGS-DINO-static$^{*}$~\cite{Gupta2024MTGS} & \FTspecific$^{\ddagger}$ & \textbf{0.094} & \textbf{0.041} & \bestang{11.0} & \textbf{0.097} & \bestang{11.0} & 0.874 & 0.858 \\
\midrule
InternVL3.5-38B~\cite{wang2025internvl3_5} & -- & 0.184 & 0.121 & \ang{26.0} & 0.250 & \ang{38.4} & 0.684 & 0.803 \\
Qwen3-VL-32B-Instruct~\cite{qwen3vltechnicalreport} 
& -- & 0.184 & 0.120 & \ang{25.3} & 0.229 & \ang{32.8} & 0.674 & 0.793 \\
Qwen3-VL-8B-Instruct~\cite{qwen3vltechnicalreport} 
& -- & 0.215 & 0.143 & \ang{30.7} & 0.249 & \ang{38.5} & 0.674 & 0.792 \\
\midrule
Qwen3-VL-8B-Instruct$^{\dagger}$ & \FTspecific & 0.110 & 0.049 & \ang{14.5} & 0.133 & \ang{19.0} & 0.787 & 0.809 \\
Qwen3-VL-8B-Instruct$^{\dagger}$ & AllD/\GazeFollowingTask & 0.112 & 0.052 & \ang{14.4} & 0.130 & \ang{18.1} & 0.786 & 0.812 \\
Qwen3-VL-8B-Instruct$^{\dagger}$ & AllD/GFo+SG & 0.114 & 0.052 & \ang{14.7} & 0.132 & \ang{19.7} & 0.812 & 0.843 \\
\bottomrule
\end{tabular}
\vspace{-4mm}
\end{table*}

\subsection{Zero-Shot Prompt Evaluation}
\label{sec:zero-shot-eval}

We evaluate different prompting strategies on gaze following and social gaze prediction using the GazeFollow and VAT test sets, respectively, with \textit{Qwen3-VL-8B-Instruct}~\cite{qwen3vltechnicalreport} 
as a representative VLM. Results are reported in Table~\ref{tab:zero-shot prompts}. Interestingly, we observe that prompting strategies exhibit markedly different behaviors across the two tasks.

For gaze following, CoT-based prompts (\VanillaCoTAcro, \StructuredCoTAcro, \CoTInContexAcro) consistently outperform the non-CoT baseline   \BaselineAcro, indicating that the task requires structured reasoning to integrate multiple visual cues. Moreover, \StructuredCoTAcro further improves performance over \VanillaCoTAcro by explicitly guiding the reasoning process. In particular, prompting the model to analyze head pose, gaze direction, and spatial relations between the subject and candidate targets leads to better localization accuracy and improved gaze direction prediction (lower angular errors). This corroborates prior work showing that structured or guided CoT is more effective than naive reasoning for visually grounded tasks \cite{chen2024vcot,ji2025spatialcot}, and that CoT improves performance on tasks involving spatial reasoning and multi-step inference \cite{lu2022learn,zhang2024multimodal}. Finally, augmenting this guidance with in-context examples (\CoTInContexAcro) provides additional gains, suggesting that exemplars help the model internalize both the reasoning process and the expected output format.

In contrast, for social gaze prediction, a simple prompt without CoT already achieves strong performance, while CoT-based variants bring limited or even negative gains. This suggests that the task relies more on high-level semantic understanding  of the scene and of the situation than on precise geometric reasoning. In this context, by benefiting from large-scale language and vision-language pretraining, VLMs are already good at capturing the global context and relational cues required for social gaze prediction, making additional reasoning scaffolding less beneficial.

\begin{figure*}[t]
\centering
\includegraphics[width=\linewidth]{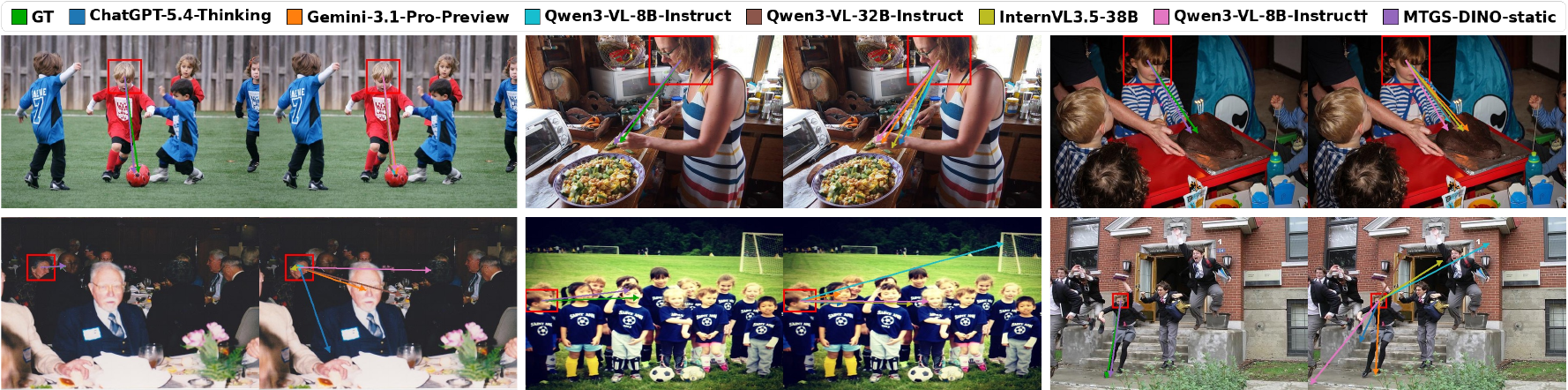}
\caption{
Qualitative results for gaze following on the GazeFollow  dataset. For each sample, the left image shows the ground-truth (GT) and MTGS-DINO-static~\cite{mtgs-static-vsgaze} predictions, while the right image presents predictions from different VLMs.
Top: successful cases.
Bottom: challenging scenarios (e.g., requiring depth understanding, involving multiple people, or targeting non-salient objects), where VLMs produce more diverse and less consistent predictions.
Zoom in for better visualization.
}

\label{fig:qualitative_gf}
\end{figure*}

\begin{table*}[t]
\caption{
Social gaze prediction results on the balanced VAT and ChildPlay test subsets.
Top block: specialized vision models, re-evaluated on the same subsets. Gaze-LLE labels and MTGS LAH/LAEO are obtained by post-processing predicted gaze points; MTGS SA uses its decoder, following~\cite{Gupta2024MTGS}.
Middle block: zero-shot VLM (with the \BaselineAcro prompt).
Bottom block: fine-tuned VLM (see Section~\ref{sec:experimentalprotocol}).
Best results are in bold, and second-best results are underlined.
}
\label{tab:vat_childplay_sg}
\centering
\footnotesize
\setlength{\tabcolsep}{4pt}
\renewcommand{\arraystretch}{1.05}
\begin{tabular}{l c c c c c c c}
\toprule
\multirow{2}{*}{Method}
& \multirow{2}{*}{FT Strategy}
& \multicolumn{3}{c}{VAT~\cite{Chong2020VideoAttentionTarget}} 
& \multicolumn{3}{c}{ChildPlay~\cite{tafasca2023childplay}} \\
\cmidrule(lr){3-5} \cmidrule(lr){6-8}
&
& F1$_{\mathrm{LAH}}\uparrow$ 
& F1$_{\mathrm{LAEO}}\uparrow$ 
& F1$_{\mathrm{SA}}\uparrow$
& F1$_{\mathrm{LAH}}\uparrow$ 
& F1$_{\mathrm{LAEO}}\uparrow$ 
& F1$_{\mathrm{SA}}\uparrow$ \\
\midrule

Gaze-LLE (ViT-B)~\cite{Ryan2025GazeLLE}
& \FTspecific
& 0.831 & \underline{0.908} & 0.732
& 0.725 & 0.692 & 0.542\\

Gaze-LLE (ViT-L)~\cite{Ryan2025GazeLLE}
& \FTspecific
& \textbf{0.833} & 0.894 & \underline{0.775}
& 0.717 & 0.794 & 0.614\\

MTGS-DINO-static~\cite{Gupta2024MTGS}
& \FTspecific$^{\ddagger}$
& \underline{0.832} & 0.895 & \textbf{0.809} 
& 0.744 & 0.862 & \underline{0.732} \\

\midrule

InternVL3.5-38B~\cite{wang2025internvl3_5}
& --
& 0.698 & 0.767 & 0.655
& 0.703 & 0.789 & 0.620 \\

Qwen3-VL-32B-Instruct~\cite{qwen3vltechnicalreport}

& --
& 0.696 & 0.826 & 0.689
& 0.662 & 0.791 & 0.574 \\

Qwen3-VL-8B-Instruct~\cite{qwen3vltechnicalreport}

& --
& 0.699 & 0.794 & 0.630
& 0.648 & 0.769 & 0.520 \\

\midrule

Qwen3-VL-8B-Instruct$^{\dagger}$
& \FTspecific
& 0.829 & \textbf{0.929} & 0.714
& \underline{0.789} & \underline{0.887} & 0.729 \\

Qwen3-VL-8B-Instruct$^{\dagger}$
& AllD/GFo+SG
& 0.808 & 0.894 & 0.707
& \textbf{0.797} & \textbf{0.905} & \textbf{0.803} \\

\bottomrule
\end{tabular}
\end{table*}

\subsection{Gaze Following Evaluation}

We evaluate three categories of methods: pure vision-based gaze models, zero-shot VLMs using the \CoTInContexAcro prompt, and fine-tuned VLMs with different training strategies. Results on GazeFollow and VAT are reported in Table~\ref{tab:sota_gf}, and results on ChildPlay (CP) are reported in Appendix~\ref{app:childplay_gf} (Table~\ref{tab:childplay_gf}) and follow the same trends.

Specialized vision-based models remain the strongest for precise gaze target localization, while zero-shot VLMs lag behind by a large margin across all datasets. Although larger zero-shot models improve localization accuracy, the gains remain limited. Fine-tuning substantially reduces this gap: compared to zero-shot \textit{Qwen3-VL-8B-Instruct}, its fine-tuned counterpart consistently lowers localization errors on GF, VAT, and CP. However, even after fine-tuning, VLMs still underperform the best specialized models in terms of L2 distance and angular error, particularly on VAT and CP. 
Joint training on all datasets (\FTfull) or adding social gaze supervision has limited impact overall (slightly worse on GF, similar on VAT, slightly better on CP). All this suggests that while VLMs contain transferable visual representations for gaze understanding, their architectures and training objectives remain suboptimal for fine-grained spatial prediction compared to dedicated gaze models.
We also explore stochastic decoding by using a non-zero temperature. We observe that outputs tend to cluster around ambiguous candidate targets, and that averaging predictions consistently improves performance as the number of samples increases. More details are provided in Appendix~\ref{sec:sampling}.

Table~\ref{tab:vlm_subset_gf_vat} further compares closed- and open-source VLMs. Gemini-3.1-Pro-Preview achieves the best performance, outperforming ChatGPT-5.4-Thinking and all open-source zero-shot models, indicating that stronger closed-source VLMs exhibit better zero-shot gaze understanding.
Nevertheless, both remain behind the specialized MTGS-DINO-static model and the fine-tuned Qwen3-VL-8B-Instruct$^{\dagger}$. This confirms that, although closed-source VLMs provide stronger performance, task-specific fine-tuning remains necessary for accurate gaze target localization.

In Fig.~\ref{fig:qualitative_gf}, when the target is salient and consistent with the activity (top), most VLMs agree with the ground truth and MTGS. In harder cases (bottom: depth reasoning, several people, non-salient targets), MTGS stays close to the annotation while VLM predictions spread over candidate targets, some implausible given the head orientation.

We also test grid-based gaze localization with VLMs to examine whether discretizing gaze target prediction alleviates the challenges of direct coordinate generation (it leads to much worse performance, Appendix~\ref{sec:grid_gaze}) and conduct ablations on model size and training data scale in fine-tuning, as detailed in Appendices~\ref{sec:ablation_model_size} and~\ref{sec:ablation_data_scale}.

\subsection{Social Gaze Prediction Evaluation}

Table~\ref{tab:vat_childplay_sg} reports social gaze results on VAT and ChildPlay for open-source VLMs compared to specialized models (MTGS and Gaze-LLE), while Table~\ref{tab:vlm_subset_gf_vat} also includes closed-source models on a subset of VAT. Overall, we observe trends similar to gaze following: zero-shot open models lag behind MTGS, larger models perform better (Qwen 8B vs 32B), and closed-source models perform significantly better, with Gemini and ChatGPT achieving comparable results.

However, in contrast to gaze following, the performance gap with specialized models is much smaller. This trend is further confirmed after fine-tuning: VLMs not only improve substantially over their zero-shot counterparts and surpass closed-source models, but also reach performance close to the specialized model. While MTGS and Gaze-LLE remain stronger on VAT, fine-tuned VLMs achieve better results on ChildPlay, particularly when trained on all datasets and combined with gaze-following supervision.
Note however that the number of social gaze samples is lower on ChildPlay than on VAT (see Table~\ref{tab:social-ann-stats} in Appendix), which may slightly affect the results. In addition, MTGS is trained on the full VSGaze, which includes additional datasets and many more negative pairs than our balanced subsets. It may also influence the performance.
Nevertheless, the overall trend suggests that VLMs can approach the performance of specialized models on social gaze tasks.

A possible explanation for the smaller gap, consistent with recent VLM literature \cite{huang2023kosmos,thumu2026socialcaptionevaluatingsocial}
, is that social gaze prediction relies more on high-level semantic and relational reasoning than on precise geometric localization. Underlying LLMs may encode knowledge about social concepts such as eye contact, shared attention, and human interactions, which can help interpret the social situation depicted in the scene. Through vision-language alignment, VLMs may leverage this knowledge, which could explain why, despite weaker gaze-following performance, they achieve comparable results on social gaze tasks. Further work is needed to validate this hypothesis by more explicitly analyzing their reasoning processes.

We also conduct blind-baseline experiments by removing image inputs while retaining the prompts and head bounding boxes. The near-chance performance indicates that visual information is essential for social gaze prediction, while head-box geometry alone provides limited cues (Appendix~\ref{app:blind_baseline}).

\begin{table*}[t]
\caption{Comparison with closed-source VLMs.  
Gaze following is evaluated on the first 1K samples of the GF 
test set, social gaze prediction on a 1K-sample subset of the VAT
test set (with 500 positive and 500 negative samples per task).
Zero-shot VLMs use the \CoTInContexAcro prompt for gaze following, 
and \BaselineAcro prompt for social gaze.
Best results are in bold, second best are underlined.
}
\label{tab:vlm_subset_gf_vat}
\centering
\footnotesize
\setlength{\tabcolsep}{3.5pt}
\renewcommand{\arraystretch}{1.05}

\begin{tabular}{l c c c c c c c c}
\toprule
\multirow{2}{*}{Method} 
& \multirow{2}{*}{FT Strategy} 
& \multicolumn{3}{c}{Gaze Following} 
& \multicolumn{3}{c}{Social Gaze Prediction} \\
\cmidrule(lr){3-5} \cmidrule(lr){6-8}
&
& Avg L2$\downarrow$ 
& Min L2$\downarrow$ 
& Ang Err$\downarrow$ 
& F1$_{\mathrm{LAH}}\uparrow$ 
& F1$_{\mathrm{LAEO}}\uparrow$ 
& F1$_{\mathrm{SA}}\uparrow$ \\
\midrule

MTGS-DINO-static~\cite{Gupta2024MTGS}
& \FTspecific$^{\ddagger}$
& \textbf{0.096} 
& \textbf{0.043} 
& \bestang{11.0} 
& \textbf{0.847} 
& \underline{0.902} 
& \textbf{0.817} \\

\midrule

ChatGPT-5.4-Thinking~\cite{openai2026gpt54thinking}
& --
& 0.164 
& 0.101 
& \ang{22.4} 
& 0.769 
& 0.879 
& 0.717 \\

Gemini-3.1-Pro-Preview~\cite{googledeepmind2026gemini31pro}
& --
& 0.131 
& 0.069 
& \ang{16.4} 
& 0.802 
& 0.850 
& \underline{0.726} \\

\addlinespace

InternVL3.5-38B~\cite{wang2025internvl3_5}
& --
& 0.183 
& 0.122 
& \ang{25.1} 
& 0.694 
& 0.767 
& 0.655 \\

Qwen3-VL-32B-Instruct~\cite{qwen3vltechnicalreport}

& --
& 0.175 
& 0.113 
& \ang{24.3} 
& 0.690 
& 0.832 
& 0.685 \\

Qwen3-VL-8B-Instruct~\cite{qwen3vltechnicalreport}

& --
& 0.211 
& 0.140 
& \ang{29.9} 
& 0.706 
& 0.804 
& 0.628 \\

\midrule

Qwen3-VL-8B-Instruct$^{\dagger}$
& \FTspecific
& \underline{0.107} 
& 0.050 
& \ang{14.2} 
& \underline{0.836}
& \textbf{0.932} 
& 0.723 \\

Qwen3-VL-8B-Instruct$^{\dagger}$
& AllD/GFo+SG
& 0.109 & \underline{0.049} & \underline{\ang{13.5}}
& 0.802 
& 0.883 
& 0.716 \\

\bottomrule
\end{tabular}
\vspace{-7mm}
\end{table*}

\section{Conclusion}

We present a benchmark of gaze understanding in VLMs, evaluating gaze following and social gaze prediction under a unified VQA formulation.
Experiments show that prompting strategies can improve zero-shot VLM performance on gaze target localization, but not on social gaze tasks, although their performance still lags behind that of specialized visual models.
Fine-tuning VLMs substantially improves performance on both tasks. However, a substantial gap remains for gaze following, suggesting limitations in precise subject-conditioned spatial grounding, whereas fine-tuned VLMs achieve performance close to that of specialized visual models on social gaze tasks.
Within the scope studied here, we show that generic vision-language pretraining alone is not sufficient for reliable gaze target estimation, and that gaze understanding remains a challenging capability for current VLMs. We hope that EyeVLM will serve as a foundation for better understanding these limitations and developing future gaze-aware VLMs.

\section*{Acknowledgement}

This research has been supported by the TUNASBE project (Towards a Unified Understanding of Attention and Social Behaviors in the Wild, grant no. 10002186) of the Swiss National Science Foundation (SNSF).

% \clearpage
{
\small
\bibliographystyle{IEEEtran}
\bibliography{neurips_2026}
}

%%%%%%%%%%%%%%%%%%%%%%%%%%%%%%%%%%%%%%%%%%%%%%%%%%%%%%%%%%%%

\newpage
\appendix

\section{Supplementary Material}
\label{app:overview}
The supplementary material is organized as follows:
\begin{itemize}
    \item \ref{sec:appendixadaptation}: implementation details of VLM fine-tuning;
    \item \ref{app:qa_examples}: QA pair examples and the joint training setup;
    \item \ref{sec:appendixdataset}: details on the datasets and data access;
    \item \ref{supp:social_gaze_sampling}: construction of the balanced social gaze subsets;
    \item \ref{app:additional_gf}: additional gaze following results on ChildPlay and cross-dataset evaluation;
    \item \ref{sec:sampling}: effect of the sampling strategy at inference;
    \item \ref{sec:ablation_model_size}: ablation on model size;
    \item \ref{sec:ablation_data_scale}: ablation on training data scale;
    \item \ref{app:parsing}: output parsing statistics;
    \item \ref{app:blind_baseline}: analysis of dataset priors with blind baselines;
    \item \ref{sec:grid_gaze}: grid-based gaze localization analysis;
    \item \ref{app:incontext_ablation}: ablation of the \CoTInContexAcro prompt settings;
    \item \ref{app:head_sensitivity}: evaluation of head localization sensitivity;
    \item \ref{app:prompts_gf}: zero-shot prompts for gaze following;
    \item \ref{app:prompts_sg}: zero-shot prompts for social gaze prediction;
    \item \ref{app:qa}: question templates used to construct QA pairs;
    \item \ref{sec:limitations}: discussion on limitations;
    \item \ref{sec:broader-impact}: discussion on broader impact.
\end{itemize}

\section{Implementation Details}
\label{sec:appendixadaptation}

We fine-tune representative Qwen-based VLMs on the constructed QA pairs using LLaMA-Factory~\cite{zheng2024llamafactory}. 
Specifically, LoRA adapters are inserted into both the self-attention projection layers
(\texttt{q\_proj}, \texttt{k\_proj}, \texttt{v\_proj}, and \texttt{o\_proj})
and the gated feed-forward projection layers
(\texttt{gate\_proj}, \texttt{up\_proj}, and \texttt{down\_proj}).
The vision encoder and language modeling head are kept frozen.
We set the LoRA rank to 8, the batch size to 64, and the learning rate to $4.0 \times 10^{-4}$.

Although Qwen3-VL models can be trained with native-resolution images, we resize all images while preserving their aspect ratios and constrain the total number of pixels to 200{,}704 ($448 \times 448$), following the resolution commonly used in prior gaze works.
The model is trained with the standard next-token prediction objective on the target JSON text.
Under this setup, gaze understanding is learned through the conventional language modeling loss over structured textual outputs.

Training is conducted on a single NVIDIA H100 GPU for 2 epochs for the gaze following task.
For social gaze prediction, we observe overfitting during training and therefore select checkpoints based on the validation sets of VAT and ChildPlay.
For fine-tuning with \FTspecific, we report the results at 500 training steps.
For fine-tuning with AllD/GFo+SG, we report the results at 5{,}000 training steps.
To improve model generalization, we randomly sample a question for each training instance at every epoch.
During inference, we uniformly use the vLLM~\cite{kwon2023efficient} framework and disable sampling for deterministic and reproducible predictions.

\section{QA Pair Examples and Joint Training}
\label{app:qa_examples}

Each training sample contains a single task-specific QA pair and no system prompt; questions or outputs from different subjects or tasks are not mixed within a sample. When an image has several annotations, it appears in several samples. Below are examples for each task (coordinates in the $[0,1000]$ Qwen convention):

\begin{tcolorbox}[breakable, colback=gray!10, colframe=gray!50, boxrule=0.5pt, arc=2pt]
\small
\textbf{In/Out.} Q: ``Is the gaze target of the person whose head is within the bounding box [333, 56, 498, 221] inside the frame?'' \\
A: \texttt{[\{"inout": 1.0\}]}

\textbf{Gaze point.} Q: ``Where is the person whose head is within the bounding box [333, 56, 498, 221] looking?'' \\
A: \texttt{[\{"gaze\_point": [140, 230]\}]}

\textbf{LAEO.} Q: ``Are the people whose heads are within the bounding boxes [313, 421, 417, 606] and [443, 185, 570, 410] in the image looking at each other?'' \\
A: \texttt{[\{"label": 1.0\}]}

\textbf{LAH.} Q: ``Is the person whose head is within the bounding box [313, 421, 417, 606] looking at the person whose head is within the bounding box [443, 185, 570, 410]?'' \\
A: \texttt{[\{"label": 1.0\}]}

\textbf{SA.} Q: ``Are the people whose heads are within the bounding boxes [313, 421, 417, 606] and [443, 185, 570, 410] in the image sharing attention?'' \\
A: \texttt{[\{"label": 0.0\}]}
\end{tcolorbox}

\noindent In the joint setting (\FTfull/GFo+SG), all gaze point, in/out, \LAH, \LAEO, and \SA QA pairs are combined into one training set. Social gaze pairs are taken from the balanced VAT and ChildPlay training subsets (Appendix~\ref{supp:social_gaze_sampling}); GazeFollow does not contribute social gaze pairs.

We do not fine-tune with Chain-of-Thought outputs, since the datasets contain no reasoning annotations. Building reliable rationales would require additional annotation or careful filtering of model-generated rationales, which could introduce noisy reasoning; in addition, CoT prompting did not improve zero-shot social gaze prediction (Table~\ref{tab:zero-shot prompts}). CoT-based fine-tuning is an interesting direction for future work.

\section{Datasets}
\label{sec:appendixdataset}

 We use three datasets, \textbf{GazeFollow}, \textbf{VideoAttentionTarget (VAT)}, and \textbf{ChildPlay}, for training and evaluation. These datasets cover complementary settings, 
 as presented below.

\mypartitle{GazeFollow \cite{Recasens2015GazeFollow}} is a large-scale benchmark for gaze following in natural images. It is collected from multiple public image sources, including SUN, MS COCO, ImageNet, and Places, and contains diverse daily scenes and activities. The dataset includes 122,143 images and 130,339 annotated person instances. Its annotations provide head 
%and eye
location information together with the corresponding 2D gaze target, making it a standard benchmark for learning gaze target localization from head cues and scene context.

\mypartitle{VideoAttentionTarget (VAT) \cite{Chong2020VideoAttentionTarget}} extends gaze following from static images to videos and emphasizes temporal modeling in dynamic scenes. It is constructed from 50 YouTube shows and contains 1,331 video clips, 71,666 frames, and 164,541 frame-level annotations. For each subject, VAT provides dense head bounding boxes and gaze point annotations, and also distinguishes whether the gaze target is inside or outside the image frame. 
Compared with GazeFollow, VAT contains less image diversity but is more suitable for evaluating temporal reasoning (whenever a temporal model is used) and in-frame versus out-of-frame gaze prediction.

\mypartitle{ChildPlay \cite{tafasca2023childplay}} is a benchmark designed for understanding gaze behavior in natural child-centered interactions. It contains 401 short videos collected from multiple online sources, comprising 120,549 frames and about 257,928 annotated instances. In addition to head bounding boxes and 2D gaze points, ChildPlay provides fine-grained gaze labels and child/adult identity annotations. Compared with conventional datasets that mainly focus on adults, ChildPlay presents more challenging interaction patterns and greater behavioral diversity, making it particularly valuable for evaluating gaze understanding in child-adult social scenarios.

\section{Social Gaze Dataset Sampling}
\label{supp:social_gaze_sampling}

We sample social gaze labels from VSGaze~\cite{Gupta2024MTGS}, focusing on the ChildPlay~\cite{tafasca2023childplay} and VideoAttentionTarget~\cite{Chong2020VideoAttentionTarget} datasets because they provide high-resolution images and cover a diverse range of social interaction scenarios, from TV interviews to child--caretaker interactions. 
First, to reduce temporal redundancy, we only retain every third frame. 
Second, for each task, we sample negative social gaze relation pairs to match the number of positive pairs, yielding a balanced set of examples. 
We further discard small head bounding boxes, as well as social gaze annotations that occur near the boundaries of gaze events, such as shared attention. These boundary cases can be ambiguous because participants may be in the middle of a gaze shift.
The resulting dataset statistics for each task and dataset are summarized in Table~\ref{tab:social-ann-stats}. Because VSGaze represents directed pairs, a pair of people may appear as both $(A,B)$ and $(B,A)$. Therefore, we also report the number of unique unordered pairs in the table for LAEO and SA.

Besides VAT and ChildPlay, VSGaze includes UCO-LAEO~\cite{MarinJimenez2019LAEONet} 
and VideoCoAtt~\cite{Fan2018SharedAttention}
. UCO-LAEO is small and only annotated for \LAEO/\LAH, which are already well covered by VAT and ChildPlay, while VideoCoAtt has low-resolution images that would confound image quality with model performance. We also balance positive and negative pairs for evaluation (Section~\ref{sec:experimentalprotocol}), so our numbers are not directly comparable with results on the original, imbalanced VSGaze test sets. Since sampling is done within the original splits and the subset indices are released, other methods can be evaluated on exactly the same benchmark.

\begin{table}[!tb]
    \caption{Number of positive and negative pairs for the social gaze tasks across VAT and ChildPlay.}
    \label{tab:social-ann-stats}
    \centering
    \footnotesize
    \setlength{\tabcolsep}{4pt}
    \renewcommand{\arraystretch}{1.05}
    \begin{tabular}{c c c c c}
        \toprule
        \multirow{2}{*}{Task} & \multicolumn{2}{c}{\textit{VAT}} & \multicolumn{2}{c}{\textit{ChildPlay}} \\
        \cmidrule(lr){2-3} \cmidrule(lr){4-5}
         & Ordered (Pos/Neg) & Unordered (Pos/Neg) & Ordered (Pos/Neg) & Unordered (Pos/Neg) \\
        \midrule
        LAH & 4736/4736 & - & 2198/2198 & -\\
        LAEO & 1874/1874 & 937/1825 & 426/426 & 213/425\\
        SA & 1958/1958 & 979/1768 & 538/538 & 269/503\\
        \bottomrule
    \end{tabular}
\end{table}

\section{Additional Gaze Following Results}
\label{app:additional_gf}

\subsection{Gaze Following Results on ChildPlay}
\label{app:childplay_gf}
Table~\ref{tab:childplay_gf} compares gaze following performance on ChildPlay. 
Zero-shot VLMs show competitive in/out prediction ability, achieving AP$_{\text{in/out}}$ and F1$_{\text{in/out}}$ scores close to specialized models, but their localization accuracy remains clearly worse, as indicated by larger L2 and angular errors. 
Fine-tuning substantially improves VLM localization performance, reducing the L2 error of Qwen3-VL-8B-Instruct from 0.245 to 0.122 and the angular error from $32.2^\circ$ to $12.7^\circ$. 
Training on all datasets and jointly with social gaze prediction brings small additional gains, with \FTfull/GFo+SG achieving the best VLM localization result.
Nevertheless, a gap remains between fine-tuned VLMs and specialized gaze models, suggesting that VLMs are strong at in/out prediction but still less accurate in precise gaze target localization.

\begin{table}[t]
\caption{Gaze following results on ChildPlay.
($^{*}$Re-evaluated. $^{\dagger}$Fine-tuned.)
}
\label{tab:childplay_gf}
\centering
\footnotesize
\setlength{\tabcolsep}{4pt}
\renewcommand{\arraystretch}{1.05}
\begin{tabular}{l c c c c c}
\toprule
Method & FT Strategy & L2$\downarrow$ & Ang Err$\downarrow$ & AP$_{\text{in/out}}$$\uparrow$ & F1$_{\text{in/out}}$$\uparrow$\\
\midrule
Gupta~\cite{gupta2022modular} & FTspec & 0.113 & -- & 0.983 & --\\
Tafasca~\cite{tafasca2023childplay} & FTspec & 0.107 & -- & 0.986 & --\\
Tafasca~\cite{Tafasca2024Sharingan} & FTspec & 0.106 & -- & 0.990 & --\\
% MTGS & 0.117 & -- & \text{0.994} & --\\
Gaze-LLE (ViT-B)~\cite{Ryan2025GazeLLE} & FTspec & 0.106 & -- & \textbf{0.994} & --\\
Gaze-LLE (ViT-L)~\cite{Ryan2025GazeLLE} & FTspec & 0.101 & -- & \textbf{0.994} & --\\
MTGS-DINO-static*~\cite{Gupta2024MTGS} & FTspec$^{\ddagger}$ & \textbf{0.099} & \bestang{9.4} & \textbf{0.994} & 0.972\\
\midrule
% LLaVA-v1.6-34B(InContext-Base) & 0.328 & $65.0^\circ$ & 0.943 & --\\
% InternVL3.5-38B-Instruct(InContext-Base) & 0.170 & $19.4^\circ$ & 0.944 & --\\
% Qwen3-VL-32B-Instruct(InContext-Base) & 0.167 & $17.6^\circ$ & 0.945 & --\\
% Qwen3-VL-8B-Instruct(InContext-Base) & 0.234 & $35.2^\circ$ & 0.945 & --\\
% \midrule
InternVL3.5-38B~\cite{wang2025internvl3_5} & -- & 0.212 & \ang{29.9} & 0.944 & 0.970\\
Qwen3-VL-32B-Instruct~\cite{qwen3vltechnicalreport} 
& -- & 0.171 & \ang{21.9} & 0.946 & 0.969\\
Qwen3-VL-8B-Instruct~\cite{qwen3vltechnicalreport} 
& -- & 0.245 & $32.2^\circ$ & 0.946 & 0.968\\
\midrule
Qwen3-VL-8B-Instruct$^{\dagger}$ & FTspec & 0.122 & \ang{12.7} & 0.961 & 0.974\\
% Qwen3-VL-8B-Instruct$^{\dagger}$ (vat+cp-GF) & 0.125 & \ang{13.7} & 0.965 & 0.966\\
% Qwen3-VL-8B-Instruct$^{\dagger}$(vat+cp-GF+SG) & 0.120 & \ang{12.8} & 0.968 & 0.966\\
Qwen3-VL-8B-Instruct$^{\dagger}$ & \FTfull/GFo & 0.118 & \ang{13.1} & 0.966 & 0.972\\
Qwen3-VL-8B-Instruct$^{\dagger}$ & \FTfull/GFo+SG  & 0.116 & \ang{12.5} & 0.969 & 0.969\\
\bottomrule
\end{tabular}
\end{table}

\subsection{Cross-Dataset Evaluation}

Table~\ref{tab:cross_dataset_vat_childplay} shows the cross-dataset performance of Qwen3-VL-8B-Instruct when fine-tuned only on GazeFollow and directly evaluated on VAT and ChildPlay. 
The model achieves reasonable L2 errors of 0.130 and 0.119, respectively. 
This suggests that GazeFollow fine-tuning enables VLMs to learn gaze localization cues that can transfer to unseen datasets. 
Nevertheless, a noticeable gap remains compared with most specialized models, indicating that current VLMs still require stronger domain adaptation or task-specific training for robust cross-dataset gaze target localization.

\begin{table}[t]
\caption{Cross-dataset results on VAT~\cite{Chong2020VideoAttentionTarget} and ChildPlay~\cite{tafasca2023childplay}. The model is only fine-tuned on GazeFollow~\cite{Recasens2015GazeFollow} dataset. *: results evaluated by us based on official code releases.}
\label{tab:cross_dataset_vat_childplay}
\centering
\footnotesize
\setlength{\tabcolsep}{4pt}
\renewcommand{\arraystretch}{1.05}
\begin{tabular}{l c c}
\toprule
\multirow{2}{*}{Method} & \multicolumn{1}{c}{VAT} & \multicolumn{1}{c}{ChildPlay} \\
\cmidrule(lr){2-2} \cmidrule(lr){3-3}
& L2$\downarrow$ & L2$\downarrow$ \\
\midrule
Chong~\cite{Chong2020VideoAttentionTarget}* & 0.119 & 0.121 \\
Jin~\cite{jin2022depth} & 0.104 & -- \\
Miao~\cite{Miao2023PDP}* & 0.109 & 0.113 \\
Gupta~\cite{gupta2022modular} & 0.137 & 0.142 \\
Tafasca~\cite{tafasca2023childplay} & 0.123 & 0.115 \\
Gaze-LLE (B)~\cite{Ryan2025GazeLLE} & 0.105 & 0.114 \\
Gaze-LLE (L)~\cite{Ryan2025GazeLLE} & \textbf{0.100} & \textbf{0.101} \\
\midrule
Qwen3-VL-8B-Instruct$^{\dagger}$ & 0.130 & 0.119 \\
\bottomrule
\end{tabular}
\end{table}

\section{Effect of Sampling Strategy}
\label{sec:sampling}

We compare Best-of-$N$ and Average-of-$N$ aggregation strategies under different numbers of samples, with the results shown in Table~\ref{tab:gazefollow_best_vs_avg_temp}.
The results show that stochastic sampling can generate better gaze predictions when multiple candidates are considered. While a single sampled output with Temp. $=0.7$ is worse than deterministic decoding, Best-of-$N$ consistently improves with larger $N$, reducing Avg L2 from 0.116 at $N=1$ to 0.055 at $N=16$. This indicates that diverse samples often contain highly accurate candidates. However, Best-of-$N$ relies on selecting the best prediction among sampled outputs and therefore reflects an upper-bound setting. In comparison, Average-of-$N$ yields only modest improvements, suggesting that averaging improves prediction stability but cannot fully exploit the best individual samples.

\begin{table}[t]
\caption{Comparison of Best-of-$N$ and Average-of-$N$ aggregation strategies on GazeFollow.
$N$ denotes the number of samples, and Temp. denotes the sampling temperature.
Best-of-$N$ selects the best sampled prediction, while Average-of-$N$ averages all sampled predictions.}
\label{tab:gazefollow_best_vs_avg_temp}
\centering
\footnotesize
\setlength{\tabcolsep}{4pt}
\renewcommand{\arraystretch}{1.05}
\begin{tabular}{cc ccc ccc}
\toprule
\multirow{2}{*}{$N$} & \multirow{2}{*}{Temp.}
& \multicolumn{3}{c}{Best-of-$N$}
& \multicolumn{3}{c}{Average-of-$N$} \\
\cmidrule(lr){3-5} \cmidrule(lr){6-8}
& 
& Avg L2$\downarrow$ & Min L2$\downarrow$ & Ang Err$\downarrow$
& Avg L2$\downarrow$ & Min L2$\downarrow$ & Ang Err$\downarrow$ \\
\midrule
1  & 0.0 & 0.110 & 0.049 & \ang{14.5} & 0.110 & 0.049 & \ang{14.5} \\
1  & 0.7 & 0.116 & 0.057 & \ang{15.1} & 0.116 & 0.057 & \ang{15.1} \\
2  & 0.7 & 0.092 & 0.044 & \ang{12.0} & 0.106 & 0.054 & \ang{13.8} \\
4  & 0.7 & 0.073 & 0.034 & \ang{9.2}  & 0.099 & 0.051 & \ang{13.0} \\
8  & 0.7 & 0.062 & 0.031 & \ang{7.8}  & 0.097 & 0.051 & \ang{12.7} \\
16 & 0.7 & \textbf{0.055} & \textbf{0.030} & \bestang{6.9} & \textbf{0.096} & \textbf{0.051} & \bestang{12.6} \\
\bottomrule
\end{tabular}
\end{table}

\section{Ablation of Model Size in Fine-tuning}
\label{sec:ablation_model_size}

Table~\ref{tab:gf_model_size} studies the effect of model size under the same fine-tuning setting. 
On GazeFollow~\cite{Recasens2015GazeFollow}, increasing the model size generally improves gaze localization, with Qwen3-VL-8B-Instruct$^{\dagger}$ achieving the best Avg L2, Min L2, and angular error. 
On cross-dataset evaluation on VAT~\cite{Chong2020VideoAttentionTarget} and ChildPlay~\cite{tafasca2023childplay}, Qwen3-VL-4B-Instruct$^{\dagger}$ obtains slightly lower L2 errors than the 8B model, with angular errors within $0.1^\circ$ on VAT and identical on ChildPlay.
These differences are very small compared with the gains from fine-tuning itself: fine-tuning the 8B model on GazeFollow only reduces its L2 error from 0.249 to 0.130 on VAT and from 0.245 to 0.119 on ChildPlay, approaching several specialized models (Table~\ref{tab:cross_dataset_vat_childplay}).
We therefore consider the cross-dataset performance of the 4B and 8B models comparable, and the results indicate that fine-tuning learns transferable gaze-target localization rather than fitting the GazeFollow distribution. Larger models fit the in-domain task slightly better, but scaling alone does not improve cross-dataset generalization in this range.

\begin{table*}[t]
\caption{Ablation on model size for fine-tuning. VAT and ChildPlay evaluations are cross dataset evaluation.}
\label{tab:gf_model_size}
\centering
\footnotesize
\setlength{\tabcolsep}{4pt}
\renewcommand{\arraystretch}{1.05}
\setlength{\tabcolsep}{5pt}
\renewcommand{\arraystretch}{1.12}
\begin{tabular}{l c c c | c c c c}
\toprule
\multirow{2}{*}{Method} & \multicolumn{3}{c}{GazeFollow~\cite{Recasens2015GazeFollow}} & \multicolumn{2}{c}{VAT~\cite{Chong2020VideoAttentionTarget}} & \multicolumn{2}{c}{ChildPlay~\cite{tafasca2023childplay}} \\
\cmidrule(lr){2-4} \cmidrule(lr){5-6} \cmidrule(lr){7-8}
& Avg L2$\downarrow$ & Min L2$\downarrow$ & Ang Err$\downarrow$ & L2$\downarrow$ & Ang Err$\downarrow$ & L2$\downarrow$ & Ang Err$\downarrow$ \\
\midrule
Qwen3-VL-2B-Instruct$^{\dagger}$ & 0.121 & 0.059 & $16.1^\circ$ & 0.131 & $19.5^\circ$ & 0.122 & $14.1^\circ$ \\
% \cdashline{1-8}
Qwen3-VL-4B-Instruct$^{\dagger}$ & 0.113 & 0.051 & $14.8^\circ$ & \textbf{0.126} & \bestang{18.2} & \textbf{0.117} & \bestang{13.1} \\
% \cdashline{1-8}
Qwen3-VL-8B-Instruct$^{\dagger}$ & \textbf{0.110} & \textbf{0.049} & \bestang{14.5} & 0.130 & $18.3^\circ$ & 0.119 & \bestang{13.1} \\
\bottomrule
\end{tabular}
\end{table*}

\section{Ablation of Training Data Scale in Fine-tuning}
\label{sec:ablation_data_scale}

Table~\ref{tab:gf_data_size} shows the impact of GazeFollow training data scale on Qwen3-VL-8B-Instruct. 
Performance improves consistently as the amount of fine-tuning data increases, with Avg L2 decreasing from 0.191 at 1k samples to 0.120 at 50k samples, and angular error dropping from \ang{25.4} to \ang{16.0}. 
Using the full training set further improves all metrics, achieving the best Avg L2 of 0.110 and angular error of \ang{14.5}. 
This demonstrates that gaze following benefits substantially from larger-scale task-specific supervision, although performance gains become more gradual at larger data scales.

\begin{table}[t]
\caption{GazeFollow test results of Qwen3-VL-8B-Instruct under different GazeFollow training sizes, compared with the mainline full-data model. Lower is better for all metrics.}
\label{tab:gf_data_size}
\centering
\footnotesize
\setlength{\tabcolsep}{4pt}
\renewcommand{\arraystretch}{1.05}
\begin{tabular}{lcccc}
\toprule
Data Scale & Avg L2$\downarrow$ & Min L2$\downarrow$ & Ang Err$\downarrow$ \\
\midrule
1k     & 0.191 & 0.120 & \ang{25.4} \\
3k     & 0.177 & 0.109 & \ang{23.9} \\
5k     & 0.155 & 0.088 & \ang{21.8} \\
10k    & 0.142 & 0.077 & \ang{20.5} \\
50k    & 0.120 & 0.058 & \ang{16.0} \\
\midrule
Full & \textbf{0.110} & \textbf{0.049} & \bestang{14.5} \\
\bottomrule
\end{tabular}
\end{table}

\section{Output Parsing Statistics}
\label{app:parsing}

Table~\ref{tab:parsing} reports the number of outputs that could not be parsed, and therefore fell back to the default values of Section~\ref{sec:experimentalprotocol}, for Qwen3-VL-8B-Instruct in the zero-shot and fine-tuned (\FTspecific) settings. Formatting failures are 0\% for all zero-shot models. A single failure occurs across all evaluations (VAT gaze point after fine-tuning, 0.0048\%). Hence the fallback has no measurable effect, and excluding non-compliant outputs instead leaves all reported metrics unchanged. This also holds for the in/out-of-frame task, whose parsing success rate is 100\%.

\begin{table}[t]
\caption{Parsing failures (failed / total test queries) for Qwen3-VL-8B-Instruct. For VAT and ChildPlay, fine-tuned gaze point evaluation is restricted to in-frame samples, following the standard protocol.}
\label{tab:parsing}
\centering
\footnotesize
\setlength{\tabcolsep}{3pt}
\begin{tabular}{l l c c c c c}
\toprule
Setting & Dataset & Gaze point & In/Out & \LAH & \LAEO & \SA \\
\midrule
\multirow{3}{*}{Zero-shot} & GF  & 0/4,782  & -- & -- & -- & -- \\
                           & VAT & 0/31,978 & 0/31,978 & 0/9,472 & 0/3,748 & 0/3,916 \\
                           & CP  & 0/22,649 & 0/22,649 & 0/4,396 & 0/852 & 0/1,076 \\
\midrule
\multirow{3}{*}{\FTspecific} & GF  & 0/4,782  & -- & -- & -- & -- \\
                             & VAT & 1/20,655 & 0/31,978 & 0/9,472 & 0/3,748 & 0/3,916 \\
                             & CP  & 0/21,390 & 0/22,649 & 0/4,396 & 0/852 & 0/1,076 \\
\bottomrule
\end{tabular}
\end{table}

\section{Analysis of Dataset Priors with Blind Baselines}
\label{app:blind_baseline}

To investigate whether the models can exploit dataset-specific priors without accessing the visual content, we evaluate two blind baselines on the same balanced VAT test set. Both baselines receive the \BaselineAcro system prompt and the head bounding-box coordinates, but no image is provided.

Specifically, we consider:
\begin{itemize}
    \item \textbf{Blind VLM}: Qwen3-VL-8B-Instruct operating in text-only mode. This setting isolates the language component of a vision-language model that has been jointly trained on vision-language data.

    \item \textbf{LLM}: Qwen3-8B~\cite{qwen3technicalreport}
, a pure-text language model trained without visual inputs.
\end{itemize}

All predictions from both blind baselines were parsed successfully. We use the same decision threshold of 0.5 for computing the F1 scores. For reference, we additionally report the zero-shot performance of Qwen3-VL-8B-Instruct with image input, denoted as \textbf{Image VLM}.

\begin{table*}[t]
    \centering
    \footnotesize
    \caption{Comparison between the image-based model and blind baselines on the balanced VAT test set. PR denotes the positive prediction rate, i.e., the proportion of samples predicted as positive.}
    \label{tab:blind_baseline}
    \begin{tabular}{lccc}
        \toprule
        Task
        & Image VLM (F1 / PR / AP)
        & Blind VLM (F1 / PR / AP)
        & LLM (F1 / PR / AP) \\
        \midrule
        LAH  & 0.699 / 0.547 / 0.661 & 0.660 / 0.902 / 0.512 & 0.026 / 0.012 / 0.503 \\
        LAEO & 0.794 / 0.668 / 0.731 & 0.690 / 0.910 / 0.533 & 0.048 / 0.058 / 0.493 \\
        SA   & 0.630 / 0.694 / 0.537 & 0.532 / 0.586 / 0.494 & 0.513 / 0.390 / 0.537 \\
        \bottomrule
    \end{tabular}
\end{table*}

As shown in Table~\ref{tab:blind_baseline}, both blind baselines achieve AP values close to 0.5 across all three tasks. Since the evaluation set is balanced, an AP of approximately 0.5 corresponds to chance-level ranking performance. This indicates that the textual prompt and relative head-box geometry alone provide limited discriminative information for identifying social gaze interactions.

Despite their near-chance AP, the two blind models exhibit substantially different prediction biases. The Blind VLM predicts positive labels for more than 90\% of the LAH and LAEO samples, resulting in relatively high F1 scores despite weak ranking performance. In contrast, the pure-text LLM predicts almost all LAH and LAEO samples as negative. For Shared Attention, both models produce more balanced positive prediction rates, while their AP values remain close to chance. These results highlight that F1 scores alone can be misleading when strong class-prediction biases are present, and further motivate reporting both AP and positive prediction rates.

Providing the image leads to clear improvements for LAH and LAEO. In particular, the Image VLM reaches an AP of 0.731 for LAEO and 0.661 for LAH, substantially outperforming both blind baselines. This confirms that the model relies on visual evidence rather than simply exploiting the prompt or spatial configuration of the detected heads. The improvement is especially pronounced for LAEO, suggesting that mutual gaze provides relatively strong visual cues that can be captured by the VLM. LAH requires more precise estimation of whether one person's gaze falls on another person's head and therefore remains more challenging.

Shared Attention remains the most difficult setting. The Image VLM obtains an AP of 0.537, which is only marginally higher than the blind baselines. This result is consistent with the additional complexity of Shared Attention, where the model must not only infer the gaze directions of multiple people but also determine whether they converge on the same external target. Such reasoning requires accurate gaze estimation together with scene-level semantic understanding, making it considerably more challenging than pairwise social gaze recognition.

\section{Grid-Based Gaze Localization Analysis}
\label{sec:grid_gaze}

To further investigate whether gaze-following performance is limited by the difficulty of generating continuous textual coordinates, we introduce a grid-based evaluation that converts gaze localization into a discrete cell-selection problem. This formulation reduces the burden of directly generating precise image coordinates and allows us to separately examine the model's spatial reasoning capability under different localization granularities.

We perform zero-shot evaluation of Qwen3-VL-8B-Instruct on the GazeFollow test set using a modified version of the \BaselineAcro prompt. The original image and textual head bounding-box information are retained. The image plane is additionally divided into a predefined $K \times K$ grid, and the model is instructed to identify the grid cell containing the predicted gaze target. Each cell is represented by a $(\text{row}, \text{column})$ pair using ordered alphabetic labels. We evaluate grid resolutions ranging from $2\times2$ to $32\times32$.

\begin{table*}[t]
    \centering
    \footnotesize
    \caption{Zero-shot grid-based gaze localization results of Qwen3-VL-8B-Instruct on the GazeFollow test set. Accuracy measures whether the predicted cell contains the ground-truth gaze target. Avg L2 denotes the normalized Euclidean distance between the predicted cell center and the annotated gaze point. The random baseline reports the exact expectation under uniform random cell selection, while the GT-cell oracle assumes perfect cell classification and reports the quantization error induced by representing each cell using its center.}
    \label{tab:grid_gaze}
    \setlength{\tabcolsep}{3pt}
    \begin{tabular}{lccc}
        \toprule
        Grid Resolution
        & Candidate Cells
        & Acc. $\uparrow$ (Model / Random)
        & Avg L2 $\downarrow$ (Model / Random / GT-Cell Oracle) \\
        \midrule
        $2\times2$   & 4    & 0.296 / 0.250 & 0.405 / 0.434 / 0.178 \\
        $4\times4$   & 16   & 0.124 / 0.063 & \textbf{0.359} / 0.461 / 0.093 \\
        $8\times8$   & 64   & 0.025 / 0.016 & 0.409 / 0.468 / 0.048 \\
        $10\times10$ & 100  & 0.007 / 0.010 & 0.464 / 0.469 / 0.038 \\
        $16\times16$ & 256  & 0.002 / 0.004 & 0.539 / 0.469 / 0.024 \\
        $32\times32$ & 1024 & 0.000 / 0.001 & 0.627 / 0.470 / 0.012 \\
        \bottomrule
    \end{tabular}
\end{table*}

All model outputs were syntactically valid. As shown in Table~\ref{tab:grid_gaze}, the model demonstrates meaningful coarse spatial localization at low grid resolutions. For $2\times2$, $4\times4$, and $8\times8$ grids, the cell-selection accuracy is consistently higher than the corresponding random baseline. In particular, the $4\times4$ configuration achieves the lowest Avg L2 error of 0.359, compared with 0.461 for random prediction, indicating that the model can identify the approximate gaze-target region substantially better than chance when the output space remains relatively compact.

However, increasing the grid resolution does not lead to progressively better localization. Although finer grids reduce the intrinsic spatial quantization error, as reflected by the monotonically decreasing GT-cell oracle error, the model's cell-selection accuracy deteriorates rapidly as the number of candidate cells grows. At $10\times10$ and finer resolutions, the model accuracy falls to or below the random baseline, while the Avg L2 error increases accordingly. The degradation becomes particularly pronounced for the $16\times16$ and $32\times32$ settings.

These results reveal a trade-off between spatial resolution and discrete selection difficulty. A coarse grid provides a relatively small output space but introduces substantial localization quantization error, whereas a fine grid offers much higher spatial precision in principle but requires the model to distinguish among a large number of candidate cells. The $4\times4$ setting achieves the best balance between these two factors in our experiments.

Importantly, the best grid-based result remains worse than the continuous text-based coordinate generation baseline. This suggests that the limited gaze-following performance cannot be attributed solely to the textual representation of continuous coordinates. Discretizing the output space does not resolve the underlying limitation and, at sufficiently fine resolutions, introduces an additional large-scale classification burden. Together, these observations indicate that the primary bottleneck lies in the model's ability to infer fine-grained gaze direction and spatial relationships from the visual input, rather than merely in the format used to express the predicted gaze location.

\section{Ablation of \CoTInContexAcro Prompt Settings}
\label{app:incontext_ablation}

Table~\ref{tab:incontext_ablation} compares different in-context example settings on GazeFollow. 
Image-text demonstrations outperform text-only demonstrations, and using two image-text examples achieves the best results across all metrics. 
This indicates that visual demonstrations provide helpful task cues for gaze localization. 

\begin{table}[t]
\caption{Ablation study on in-context example settings for Qwen3-VL-32B-Instruct on GazeFollow. 
T denotes text-only examples, while I+T denotes image-text examples.}
\label{tab:incontext_ablation}
\centering
\footnotesize
\setlength{\tabcolsep}{5pt}
\renewcommand{\arraystretch}{1.1}
\begin{tabular}{l c c c c}
\toprule
Modality & \# Examples & Avg L2$\downarrow$ & Min L2$\downarrow$ & Ang Err$\downarrow$ \\
\midrule
T   & 2 & 0.193 & 0.127 & \ang{27.3} \\
I+T & 1 & 0.189 & 0.125 & \ang{26.2} \\
I+T & 2 & \textbf{0.184} & \textbf{0.120} & \bestang{25.3} \\
\bottomrule
\end{tabular}
\end{table}

\section{Head Localization Sensitivity Evaluation}
\label{app:head_sensitivity}

One of the key assumptions in this work is that VLMs are capable of understanding head localization inputs. In this section, we verify this assumption by benchmarking the ability of VLMs to identify a person's head given a bounding box as input. The objective is to construct a balanced dataset of positive bounding boxes containing a person's head and negative bounding boxes containing none, then evaluate prompting strategies for head presence detection.

\begin{figure}[t]
    \centering
    \begin{subfigure}[t]{0.47\linewidth}
        \centering
        \includegraphics[width=\linewidth]{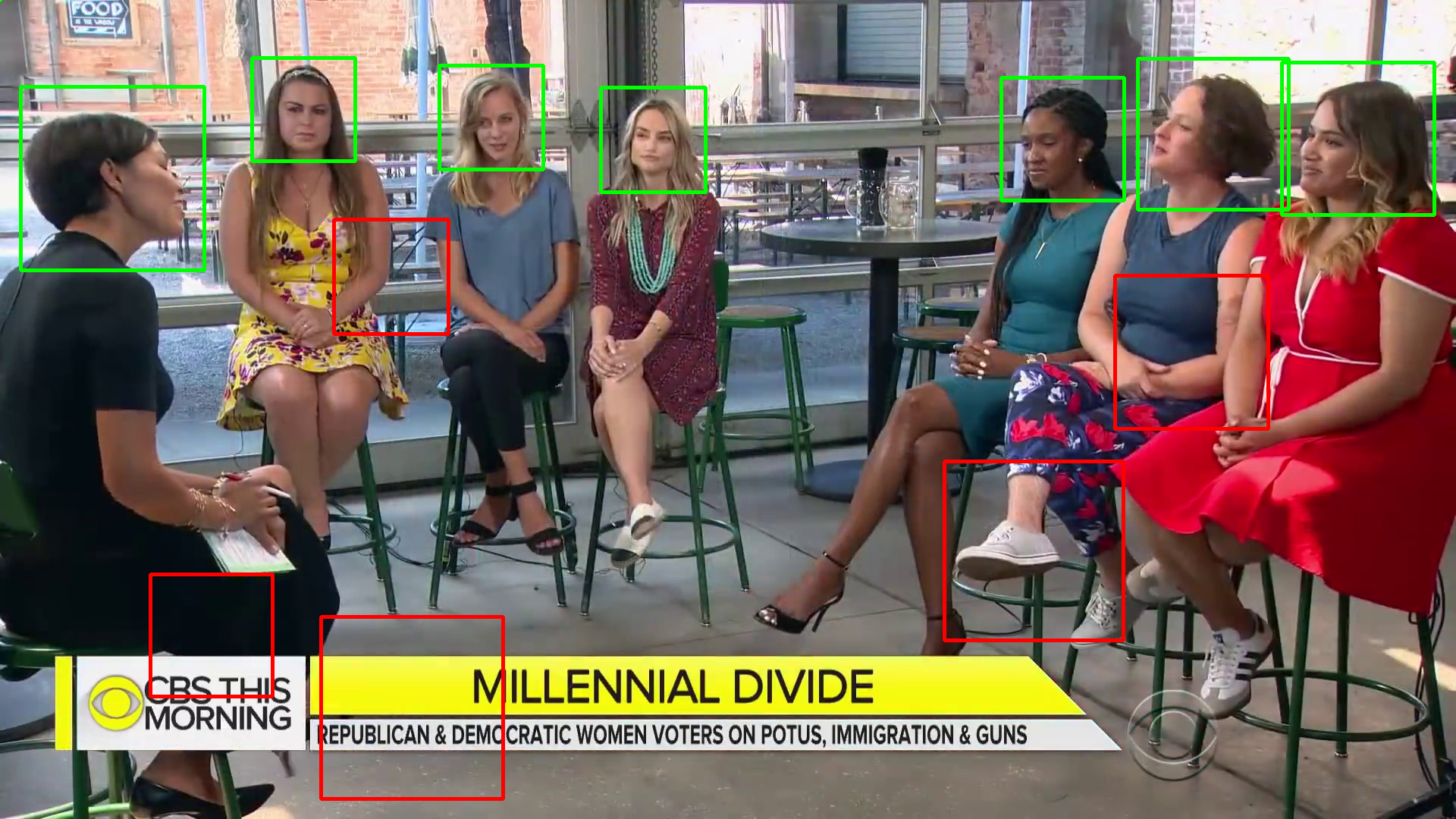}
        \caption{Examples of positive (green) and negative (red) head bounding boxes generated from the VAT test set. Negative boxes are constrained to have no overlap with any annotated head.}
        \label{fig:generated_head}
    \end{subfigure}
    \hfill
    \begin{subfigure}[t]{0.50\linewidth}
        \centering
        \includegraphics[width=\linewidth]{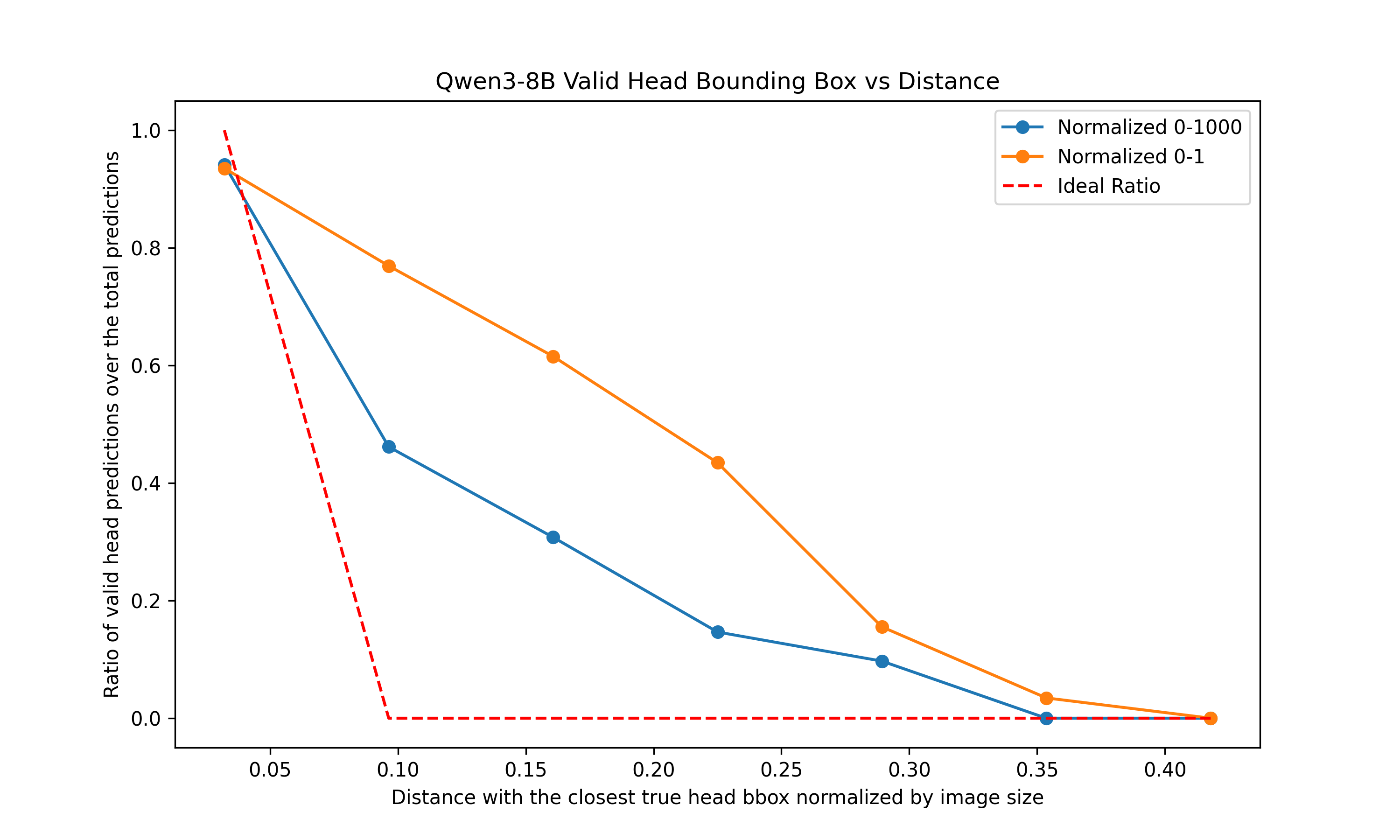}
        \caption{Analysis of positive prediction given bounding box distance to the closest ground truth. Positive boxes have a distance of 0 and negative > 0.}
        \label{fig:plot_head_loc}
    \end{subfigure}

\end{figure}

\noindent \textbf{Dataset.}
We manually selected a subset of 189 images from the VAT test set, drawn from different videos and excluding images in which heads are unannotated (e.g., small heads in the background). For each annotated head in each image, a corresponding negative bounding box is generated with no overlap with any ground-truth head. Figure~\ref{fig:generated_head} illustrates both sets: ground-truth positive boxes are shown in green and generated negative boxes in blue. The resulting dataset comprises 958 question-answering pairs, balanced with 50\% positive and 50\% negative examples.

\noindent \textbf{Settings.}
We evaluate the \textit{Qwen3-VL-8B-Instruct} model on this task. All images are resized so that the longest side is 448 pixels while preserving the original aspect ratio. We compare two bounding box input representations: coordinates normalized to the range $[0, 1000]$ and coordinates normalized to the range $[0, 1]$. In both cases, the model is queried with the following prompt:

\begin{tcolorbox}[
  colback=gray!10,   % 背景浅灰
  colframe=gray!50,  % 边框灰色
  boxrule=0.5pt,
  arc=4pt,           % 圆角
  left=6pt,right=6pt,top=6pt,bottom=6pt,
  breakable
]

\begin{Verbatim}[
  breaklines=true,
  breakanywhere=true,
  breaksymbolleft={},
  breaksymbolright={}
]
Look at the region defined by the bounding box coordinate <bounding box coordinates>. Is there a person's head inside this bounding box? Answer only `Yes' or `No'.
\end{Verbatim}
\end{tcolorbox}

\noindent \textbf{Results.}
We report accuracy, sensitivity (recall, $TP/(TP+FN)$), precision, and F1. Sensitivity is the primary metric of interest, since all experiments in the main paper query the model only with valid head bounding boxes; precision and F1 additionally account for false positives on negative boxes.\\
As shown in Table~\ref{tab:normalization_bbox_ablation}, with $[0,1000]$ coordinates the model achieves a sensitivity of 0.94, a precision of 0.84, and an F1 of 0.89, demonstrating a strong ability to recognize the presence of a head when a positive bounding box is provided. However, accuracy drops to 0.88, revealing a tendency to predict ``yes'' even for negative boxes. This suggests that the model does not precisely localize bounding box coordinates, but rather makes predictions based on approximate spatial reasoning. This is further supported by the distance analysis in Figure~\ref{fig:plot_head_loc}, which shows that false positives on negative boxes tend to occur when those boxes are in close proximity to a ground-truth head.\\
Overall, these results confirm that Qwen3 can approximately localize bounding box coordinates and reliably identify a person's head when a positive bounding box is provided as input, validating the core assumption of our approach.

\begin{table}[t]
\caption{Head presence detection with Qwen3-VL-8B-Instruct on 958 balanced bounding-box queries, under different input coordinate normalizations.}
\label{tab:normalization_bbox_ablation}
\centering
\footnotesize
\setlength{\tabcolsep}{8pt}
\renewcommand{\arraystretch}{1.1}
\begin{tabular}{lcccc}
\toprule
Setting & Accuracy$\uparrow$ & Sensitivity$\uparrow$ & Precision$\uparrow$ & F1$\uparrow$ \\
\midrule
Normalized 0--1    & 0.77 & 0.94 & 0.70 & 0.80 \\
Normalized 0--1000 & 0.88 & 0.94 & 0.84 & 0.89 \\
\bottomrule
\end{tabular}
%  TODO(authors): precision/F1 for the 0--1 row are derived from accuracy and sensitivity on the balanced set (spec. = 2*acc - sens); please confirm with the raw counts.
\end{table}

\vspace{-4mm}
\section{Zero-Shot Prompts for \GazeFollowingTask}
\label{app:prompts_gf}

For simplicity, we use \texttt{<Task description>} to denote the following description of the gaze following task.

\begin{tcolorbox}[
  colback=gray!10,   % 背景浅灰
  colframe=gray!50,  % 边框灰色
  boxrule=0.5pt,
  arc=4pt,           % 圆角
  left=6pt,right=6pt,top=6pt,bottom=6pt,
  breakable
]

\begin{Verbatim}[
  breaklines=true,
  breakanywhere=true,
  breaksymbolleft={},
  breaksymbolright={}
]
You are a vision assistant specializing in **human gaze analysis**. Your task is to analyze an input image and determine **where a person of interest is looking at**. You must output two things:

- **inout**: a float number in [0,1] estimating how likely the person's gaze target lies within the current image area.
- **gaze_point**: the (x,y) image coordinates of the person's point of attention (that is their gaze point). The origin (0,0) corresponds to the top-left corner. x increases to the right, y increases downward.

Note that you **must always provide a gaze point, even if you believe the person is looking outside the image**.

The person of interest is identified by the **bounding box coordinates** of their head in the user's input.
\end{Verbatim}
\end{tcolorbox}

\subsection{\BaselineAcro}

\begin{tcolorbox}[
  colback=gray!10,   % 背景浅灰
  colframe=gray!50,  % 边框灰色
  boxrule=0.5pt,
  arc=4pt,           % 圆角
  breakable,
  left=6pt,right=6pt,top=6pt,bottom=6pt
]

\begin{Verbatim}[
  breaklines=true,
  breakanywhere=true,
  breaksymbolleft={},
  breaksymbolright={}
]
<Task description>

### Input/Output Format:

- **Input:** An image followed by a set of **bounding box coordinates** of the head of the person of interest.
- **Output:** Report in JSON format:
  - **inout**: the probability of looking inside the frame;
  - **gaze_point** the estimated gaze point coordinates.

**Required Output Format:**
### Gaze Point ###
```json
[
{"inout": <0.0-1.0>,"gaze_point": [x, y]}
]
```
\end{Verbatim}
\end{tcolorbox}

\subsection{\VanillaCoTAcro}

\begin{tcolorbox}[
  colback=gray!10,   % 背景浅灰
  colframe=gray!50,  % 边框灰色
  boxrule=0.5pt,
  arc=4pt,           % 圆角
  breakable,
  left=6pt,right=6pt,top=6pt,bottom=6pt
]

\begin{Verbatim}[
  breaklines=true,
  breakanywhere=true,
  breaksymbolleft={},
  breaksymbolright={}
]
<Task description>

**In order to achieve your task, you can first do a step-by-step analysis.**

### Input/Output Format:
- **Input:** An image followed by a set of **bounding box coordinates** of the head of the person of interest.
- **Output:** Report in JSON format:
  - **inout**: the probability of looking inside the frame;
  - **gaze_point** the estimated gaze point coordinates.

**Required Output Format:**
### Reasoning ###
<Your step-by-step reasoning here>

### Gaze Point ###
```json
[
{"inout": <0.0-1.0>, "gaze_point": [x, y]}
]
```
\end{Verbatim}
\end{tcolorbox}

\subsection{\StructuredCoTAcro}
\begin{tcolorbox}[
  colback=gray!10,   % 背景浅灰
  colframe=gray!50,  % 边框灰色
  boxrule=0.5pt,
  arc=4pt,           % 圆角
  breakable,
  left=6pt,right=6pt,top=6pt,bottom=6pt
]

\begin{Verbatim}[
  breaklines=true,
  breakanywhere=true,
  breaksymbolleft={},
  breaksymbolright={}
]
<Task description>

**In order to achieve your task, you can first do a step-by-step analysis**. Your entire response, including reasoning and final answer, must not exceed 1000 tokens and avoid repetition.

For instance, you can do the following:

1. Describe what is happening in the image.
2. Identify the activity or activities in which the person of interest is engaged in.
3. Determine the direction of gaze of the person of interest. Analyze the orientation of their head and eyes (if visible) to infer their gaze direction.
4. Identify and locate all plausible gaze targets (objects and/or persons) in the image that are visible to the person of interest, that is, that are within their field-of-view. Evaluate as well whether the person of interest is looking at something that is within the image area, or outside the image area.
5. Describe the spatial relationships between the person of interest and each of these plausible gaze targets.
6. Based on the scene context and information above, estimate how possible the person of interest is looking inside the frame and which plausible gaze target is the person of interest most likely looking at regardless of whether it is inside or outside the frame.
7. Using your reasoning, localize the most probable target of attention of the person of interest as a point. Use this point as the final gaze point. Provide the requested outputs (inout and gaze_point).

### Input/Output Format:
- **Input:** An image followed by a set of **bounding box coordinates** of the head of the person of interest.
- **Output:** Report in JSON format:
  - **inout**: the probability of looking inside the frame;
  - **gaze_point** the estimated gaze point coordinates.

**Required Output Format:**
### Reasoning ###
<Your step-by-step reasoning here>

### Gaze Point ###
```json
[
{"inout": <0.0-1.0>, "gaze_point": [x, y]}
]
```
\end{Verbatim}
\end{tcolorbox}

% \subsection{\VanillaInContextAcro}

% \begin{tcolorbox}[
%   colback=gray!10,   % 背景浅灰
%   colframe=gray!50,  % 边框灰色
%   boxrule=0.5pt,
%   arc=4pt,           % 圆角
%   breakable,
%   left=6pt,right=6pt,top=6pt,bottom=6pt
% ]

% \begin{Verbatim}[
%   breaklines=true,
%   breakanywhere=true,
%   breaksymbolleft={},
%   breaksymbolright={}
% ]
% user:
% <image>Where is the focus of the subject whose head is bounded by [0.183, 0.505, 0.319, 0.687]

% assistant:
% [
% {"inout": 1.0, "gaze_point": [0.27, 0.83]}
% ]

% user:
% <image>What is the gaze point of the individual located at [0.353, 0.152, 0.490, 0.281]?

% assistant:
% [
% {"inout": 1.0, "gaze_point": [0.19, 0.28]}
% ]

% user:
% Examples provided. Answer the next query in the same JSON format.
% \end{Verbatim}
% \end{tcolorbox}

\subsection{\CoTInContexAcro}

\begin{tcolorbox}[
  colback=gray!10,   % 背景浅灰
  colframe=gray!50,  % 边框灰色
  boxrule=0.5pt,
  arc=4pt,           % 圆角
  left=6pt,right=6pt,top=6pt,bottom=6pt,
  breakable
]

\begin{Verbatim}[
  breaklines=true,
  breakanywhere=true,
  breaksymbolleft={},
  breaksymbolright={}
]
<Task description>

**In order to achieve your task, you can first do a step-by-step analysis.** Your entire response, including reasoning and final answer, must not exceed 1000 tokens and avoid repetition.

### Input/Output Format:
- **Input:** An image followed by a set of **bounding box coordinates** of the head of the person of interest.
- **Output:** Report in JSON format:
  - **inout**: the probability of looking inside the frame;
  - **gaze_point** the estimated gaze point coordinates.

**Required Output Format:**
### Reasoning ###
<Your step-by-step reasoning here>

### Gaze Point ###
```json
[
{"inout": <0.0-1.0>, "gaze_point": [x, y]}
]
```

To help you understand how to do the analysis, I provide you with two examples:

user:
<image>[0.0, 0.013, 0.279, 0.475]

assistant:
### Reasoning ###
1. The image shows a café or casual dining setting with two people sitting at a table. The person of interest is the woman on the left, identified by the provided head bounding box.
2. Her head and upper body are turned toward the right side of the image, which indicates that her attention is directed toward the man beside her or toward an object near him.
3. Her visible facial orientation suggests a rightward gaze rather than a downward look at the table or an upward look toward the background shelves. This makes the interaction area on the right side of the image the most likely region of attention.
4. Within that rightward region, the most plausible gaze targets are the man’s face and the large white cup he is holding. The cup is the stronger candidate because it is visually salient, close to her line of sight, and central to the man’s ongoing action, while his face is partly occluded by the cup. By contrast, other objects such as the glass on the table or the background shelves are less likely because they are either lower than her apparent line of sight or less relevant to the immediate interaction.
5. Therefore, the most likely gaze target is the large white cup on the right side of the image. Since this target is clearly inside the frame, the probability of looking inside the image should be high.

### Gaze Point ###
```json
[
{"inout": 1.0, "gaze_point": [0.81, 0.24]}
]
```

user:
<image>[0.333, 0.056, 0.498, 0.221]

assistant:
### Reasoning ###
1. The image shows a field sport scene with several players. The person of interest is the player in red at the center, identified by the given head bounding box.
2. His body is moving forward, but his head is turned toward the left side of the image. His face orientation suggests that his attention is not on the ground directly below him or on the distant players in the background.
3. The most visually salient object in the direction of his head and eyes is the small ball on the left side of the image. It is close to him, in front of his hurley, and consistent with the ongoing action.
4. Other possible targets, such as the nearby teammate behind him or the player in blue on the right, are less likely because they are not aligned with his apparent line of sight. The ball is much more consistent with both the sports context and his immediate action.
5. The ball is clearly inside the frame, so the probability that he is looking inside the image is very high.
6. Therefore, the most plausible gaze target is the ball on the left side of the image, with the gaze point placed near its center.

### Gaze Point ###
```json
[
{"inout": 1.0, "gaze_point": [0.14, 0.23]}
]
```

Example provided. Now, analyze the following image.
\end{Verbatim}
\end{tcolorbox}

\section{Zero-Shot Prompts for SG}
\label{app:prompts_sg}

For simplicity, we use \texttt{<Task description>} to denote the description of the social gaze prediction task.

For LAEO:

\begin{tcolorbox}[
  colback=gray!10,   % 背景浅灰
  colframe=gray!50,  % 边框灰色
  boxrule=0.5pt,
  arc=4pt,           % 圆角
  breakable,
  left=6pt,right=6pt,top=6pt,bottom=6pt
]

\begin{Verbatim}[
  breaklines=true,
  breakanywhere=true,
  breaksymbolleft={},
  breaksymbolright={}
]
You are a helpful assistant specializing in **human gaze analysis**. Your task is to analyze an input image, and determine **whether the people of interest are looking at each other**.

The people of interest are identified by the **bounding box coordinates** of their heads in the user's input.
\end{Verbatim}
\end{tcolorbox}

For LAH:

\begin{tcolorbox}[
  colback=gray!10,   % 背景浅灰
  colframe=gray!50,  % 边框灰色
  boxrule=0.5pt,
  arc=4pt,           % 圆角
  breakable,
  left=6pt,right=6pt,top=6pt,bottom=6pt
]

\begin{Verbatim}[
  breaklines=true,
  breakanywhere=true,
  breaksymbolleft={},
  breaksymbolright={}
]
You are a helpful assistant specializing in **human gaze analysis**. Your task is to analyze an input image, and determine **whether person A is looking at person B**. 

The people of interest are identified by the **bounding box coordinates** of their heads in the user's input.
\end{Verbatim}
\end{tcolorbox}

For SA:

\begin{tcolorbox}[
  colback=gray!10,   % 背景浅灰
  colframe=gray!50,  % 边框灰色
  boxrule=0.5pt,
  arc=4pt,           % 圆角
  breakable,
  left=6pt,right=6pt,top=6pt,bottom=6pt
]

\begin{Verbatim}[
  breaklines=true,
  breakanywhere=true,
  breaksymbolleft={},
  breaksymbolright={}
]
You are a helpful assistant specializing in **human gaze analysis**. Your task is to analyze an input image, and determine **whether the people of interest are sharing attention, i.e., looking towards the same object or person in the scene.**. 

The people of interest are identified by the **bounding box coordinates** of their heads in the user's input.
\end{Verbatim}
\end{tcolorbox}

In addition, we use \texttt{<task>} to denote the social gaze task.

For LAEO:

\begin{tcolorbox}[
  colback=gray!10,   % 背景浅灰
  colframe=gray!50,  % 边框灰色
  boxrule=0.5pt,
  arc=4pt,           % 圆角
  breakable,
  left=6pt,right=6pt,top=6pt,bottom=6pt
]

\begin{Verbatim}[
  breaklines=true,
  breakanywhere=true,
  breaksymbolleft={},
  breaksymbolright={}
]
looking at each other
\end{Verbatim}
\end{tcolorbox}

For LAH:

\begin{tcolorbox}[
  colback=gray!10,   % 背景浅灰
  colframe=gray!50,  % 边框灰色
  boxrule=0.5pt,
  arc=4pt,           % 圆角
  breakable,
  left=6pt,right=6pt,top=6pt,bottom=6pt
]

\begin{Verbatim}[
  breaklines=true,
  breakanywhere=true,
  breaksymbolleft={},
  breaksymbolright={}
]
person A looking at person B
\end{Verbatim}
\end{tcolorbox}

For SA:

\begin{tcolorbox}[
  colback=gray!10,   % 背景浅灰
  colframe=gray!50,  % 边框灰色
  boxrule=0.5pt,
  arc=4pt,           % 圆角
  breakable,
  left=6pt,right=6pt,top=6pt,bottom=6pt
]

\begin{Verbatim}[
  breaklines=true,
  breakanywhere=true,
  breaksymbolleft={},
  breaksymbolright={}
]
shared attention
\end{Verbatim}
\end{tcolorbox}

\subsection{\BaselineAcro}

\begin{tcolorbox}[
  colback=gray!10,   % 背景浅灰
  colframe=gray!50,  % 边框灰色
  boxrule=0.5pt,
  arc=4pt,           % 圆角
  breakable,
  left=6pt,right=6pt,top=6pt,bottom=6pt
]

\begin{Verbatim}[
  breaklines=true,
  breakanywhere=true,
  breaksymbolleft={},
  breaksymbolright={}
]
<Task description>

### Input/Output Format:
- **Input:** An image followed by a pair of **bounding box coordinates** of the heads of the people of interest.
- **Output:** Report in JSON format:
  - **label** the probability of <task>.

**Required Output Format:**
### Social Gaze Label ###
```json
[
{"label": <0.0-1.0>}}
]
```
\end{Verbatim}
\end{tcolorbox}

\subsection{\VanillaCoTAcro}

\begin{tcolorbox}[
  colback=gray!10,   % 背景浅灰
  colframe=gray!50,  % 边框灰色
  boxrule=0.5pt,
  arc=4pt,           % 圆角
  breakable,
  left=6pt,right=6pt,top=6pt,bottom=6pt
]

\begin{Verbatim}[
  breaklines=true,
  breakanywhere=true,
  breaksymbolleft={},
  breaksymbolright={}
]
<Task description>

**In order to achieve your task, you can first do a step-by-step analysis**.

### Input/Output Format:
- **Input:** An image followed by a pair of **bounding box coordinates** of the heads of the people of interest.
- **Output:** Report in JSON format:
  - **label** the probability of <task>.

**Required Output Format:**
### Reasoning ###
<Your step-by-step reasoning here>

### Social Gaze Label ###
```json
[
{"label": <0.0-1.0>} 
]
```
\end{Verbatim}
\end{tcolorbox}

\subsection{\StructuredCoTAcro}

\begin{tcolorbox}[
  colback=gray!10,   % 背景浅灰
  colframe=gray!50,  % 边框灰色
  boxrule=0.5pt,
  arc=4pt,           % 圆角
  breakable,
  left=6pt,right=6pt,top=6pt,bottom=6pt
]

\begin{Verbatim}[
  breaklines=true,
  breakanywhere=true,
  breaksymbolleft={},
  breaksymbolright={}
]
<Task description>

**In order to achieve your task, you can first do a step-by-step analysis**. For instance, you can do the following:

1. Describe what is happening in the image.
2. Identify the activity or activities in which the people of interest are engaged in.
3. Determine the direction of gaze of the people of interest. Analyze the orientation of their head and eyes (if visible) to infer their gaze direction.
4. Identify and locate all plausible gaze targets (objects and/or persons) in the image that are visible to the people of interest, that is, that are within their field-of-view.
5. Describe the spatial relationships between the people of interest and each of these plausible gaze targets.
6. Based on the scene context and information above, determine whether the people of interest are engaged in social gaze.
7. Using your reasoning, estimate the probability of <task> for the people of interest. 

### Input/Output Format:
- **Input:** An image followed by a pair of **bounding box coordinates** of the heads of the people of interest.
- **Output:** Report in JSON format:
  - **label** the probability of <task>.

**Required Output Format:**
### Reasoning ###
<Your step-by-step reasoning here>

### Social Gaze Label ###
```json
[
{"label": <0.0-1.0>} 
]
```
\end{Verbatim}
\end{tcolorbox}

\subsection{\CoTInContexAcro}

We provide the in-context prompt for VAT LAEO as an example. The in-context prompts for the other tasks and datasets follow the same format, and are available in the codebase.

\begin{tcolorbox}[
  colback=gray!10,   % 背景浅灰
  colframe=gray!50,  % 边框灰色
  boxrule=0.5pt,
  arc=4pt,           % 圆角
  breakable,
  left=6pt,right=6pt,top=6pt,bottom=6pt
]

\begin{Verbatim}[
  breaklines=true,
  breakanywhere=true,
  breaksymbolleft={},
  breaksymbolright={}
]
<Task description>

**In order to achieve your task, you can first do a step-by-step analysis**. Your entire response, including reasoning and final answer, must not exceed 1000 tokens and avoid repetition.

### Input/Output Format:
- **Input:** An image followed by a pair of **bounding box coordinates** of the heads of the people of interest.
- **Output:** Report in JSON format:
  - **label** the probability of looking at each other.

**Required Output Format:**
### Reasoning ###
<Your step-by-step reasoning here>

### Social Gaze Label ###
```json
[
{"label": <0.0-1.0>}
]
```

To help you understand how to do the analysis, I provide you with two examples:

user:
<image> Analyze the image and determine whether the people with head boxes at [0.313, 0.421, 0.417, 0.606] and [0.443, 0.185, 0.570, 0.410] are looking at each other.

assistant:
### Reasoning ###
1. The two people are a woman in the center and a child to her lower left, sitting together on a bed with others around them.
2. They appear to be interacting in a group conversation or family scene.
3. The child’s head is tilted upward toward the woman, so the child is looking at the woman’s face/head.
4. The woman’s head is turned slightly down and left toward the child, indicating her gaze is directed at the child.
5. Other plausible gaze targets exist in the room, but both people’s faces are oriented toward each other more than toward anyone else.
6. So the people are looking at each other.

### Social Gaze Label ###
```json
[
  {"label": 1}
]
```

user:
<image> Analyze the image and determine whether the people with head boxes at [0.330, 0.210, 0.407, 0.348] and [0.427, 0.294, 0.527, 0.470] are looking at each other.

assistant:
### Reasoning ###
1. The two people are soldiers seated among others while handling gear on the ground.
2. The scene is a busy military prep area, and both appear focused on equipment rather than on each other.
3. The left soldier has head angled downward and slightly right, toward the items in front of him.
4. The right soldier also has head angled downward, looking at the gear or his hands.
5. Although they are near each other, their gaze directions do not align with each other’s heads.
6. So they are not looking at each other.

### Social Gaze Label ###
```json
[
  {"label": 0}
]
```

Examples provided. Now, analyze the following image.
\end{Verbatim}
\end{tcolorbox}

\section{Question Templates in QA Pair Construction}
\label{app:qa}

% \begin{tcolorbox}[colback=gray!10, colframe=gray!50, boxrule=0.5pt, arc=2pt]
\begin{tcolorbox}[
    breakable,
    colback=gray!10,
    colframe=gray!50,
    boxrule=0.5pt,
    arc=2pt
]
\textbf{Gaze Following In/Out Questions}
\begin{itemize}
    \item ``Is the gaze target of <OnePerson> inside the frame?''
    \item ``Is <OnePerson> looking at something in the scene?''
    \item ``What is the probability that the focus of <OnePerson> lies inside the current image area?''
    \item ``How probable is it that <OnePerson> is looking at something within the boundaries of the scene?''
    \item ``How certain are you that what <OnePerson> is looking at is shown in the image?''
    \item ``Estimate the probability that the gaze point of <OnePerson> lies within the frame.''
    \item ``Rate the likelihood that the gaze of <OnePerson> is directed at something within the frame.''
    \item ``Assign a probability representing the chance that the visual target of <OnePerson> lies inside the frame.''
    \item ``Determine the probability that the focus of <OnePerson> is contained within the scene.''
    \item ``Evaluate the chance that <OnePerson> is looking at something that is in this photograph.''
\end{itemize}

\textbf{Gaze Following Gaze Point Questions}
\begin{itemize}
    \item ``Where is <OnePerson> looking?''
    \item ``What is the gaze point of <OnePerson>?''
    \item ``Where is the focus of <OnePerson>?''
    \item ``Estimate the gaze point of <OnePerson>.''
    \item ``Localize the gaze point of <OnePerson> within the image.''
    \item ``Where is the visual target of <OnePerson> in the image?''
    \item ``Where is the attention of <OnePerson> currently directed?''
    \item ``Find the point of interest for <OnePerson>.''
    \item ``Determine the gaze destination of <OnePerson>.''
    \item ``What are the coordinates of the gaze target for <OnePerson> within the frame?''
\end{itemize}

\textbf{LAEO Questions}
\begin{itemize}
\item ``Are the <TwoPersons> in the image looking at each other?''
\item ``Is there eye contact between the <TwoPersons>?''
\item ``There are <TwoPersons> in the image. Do these two people appear to be making eye contact?''
\item ``We can see <TwoPersons> in the image. Are their gazes directed toward one another?''
\item ``Are the <TwoPersons> mutually looking at each other?''
\item ``Estimate the probability that the <TwoPersons> are engaged in eye contact.''
\item ``What is the chance that <OnePerson> and <OnePerson> are looking at each other?''
\item ``Is there eye contact between <OnePerson> and <OnePerson>?''
\item ``Do the <TwoPersons> appear to be looking at each other?''
\item ``Would you say the <TwoPersons> are making eye contact?''
\end{itemize}

\textbf{LAH Questions}
\begin{itemize}
\item ``Is <OnePerson> looking at <OnePerson>?''
\item ``There are <TwoPersons> in the image. Does <OnePerson> appear to be looking at <OnePerson>?''
\item ``We can see <TwoPersons> in the image. Is <OnePerson> looking at <OnePerson>?''
\item ``Estimate the probability that <OnePerson> is looking at <OnePerson>?''
\item ``What is the chance that <OnePerson> is looking at <OnePerson>?''
\item ``Does <OnePerson> seem to be looking toward <OnePerson>?''
\item ``How likely is it that <OnePerson> is looking at <OnePerson>?''
\item ``Give a probability that <OnePerson> is looking at <OnePerson>.''
\item ``There are <TwoPersons> visible in the image. Does <OnePerson> seem to be looking at <OnePerson>?''
\item ``Would you say <OnePerson> is looking at <OnePerson>?''
\end{itemize}

\textbf{SA Questions}

Same format as LAEO questions: replace “eye contact” and “looking at each other” with “sharing attention” to obtain the questions.

\medskip

\textbf{\texttt{<OnePerson>}}
\begin{itemize}
    \item ``the <person> whose head is within the bounding box $[x_{\min},\, y_{\min},\, x_{\max},\, y_{\max}$] ''
    \item ``the <person> located at $[x_{\min},\, y_{\min},\, x_{\max},\, y_{\max}$] ''
    \item ``the <person> whose head is enclosed by the bounding box $[x_{\min},\, y_{\min},\, x_{\max},\, y_{\max}$]''
    \item ``the <person> whose head is bounded by $[x_{\min},\, y_{\min},\, x_{\max},\, y_{\max}$]''
    \item ``the <person> whose head is identified in the region $[x_{\min},\, y_{\min},\, x_{\max},\, y_{\max}$]''
\end{itemize}

\textbf{\texttt{<TwoPersons>}}
\begin{itemize}
    \item "the <persons> whose heads are within the bounding boxes $[x_{\min}^{(1)},\, y_{\min}^{(1)},\, x_{\max}^{(1)},\, y_{\max}^{(1)}]$ and $[x_{\min}^{(2)},\, y_{\min}^{(2)},\, x_{\max}^{(2)},\, y_{\max}^{(2)}]$"
    \item "the <persons> located at $[x_{\min}^{(1)},\, y_{\min}^{(1)},\, x_{\max}^{(1)},\, y_{\max}^{(1)}]$ and $[x_{\min}^{(2)},\, y_{\min}^{(2)},\, x_{\max}^{(2)},\, y_{\max}^{(2)}]$"
    \item "the <persons> whose heads are enclosed by the bounding boxes $[x_{\min}^{(1)},\, y_{\min}^{(1)},\, x_{\max}^{(1)},\, y_{\max}^{(1)}]$ and $[x_{\min}^{(2)},\, y_{\min}^{(2)},\, x_{\max}^{(2)},\, y_{\max}^{(2)}]$"
    \item "the <persons> whose heads are bounded by $[x_{\min}^{(1)},\, y_{\min}^{(1)},\, x_{\max}^{(1)},\, y_{\max}^{(1)}]$ and $[x_{\min}^{(2)},\, y_{\min}^{(2)},\, x_{\max}^{(2)},\, y_{\max}^{(2)}]$"
    \item "the <persons> whose heads are identified in the regions $[x_{\min}^{(1)},\, y_{\min}^{(1)},\, x_{\max}^{(1)},\, y_{\max}^{(1)}]$ and $[x_{\min}^{(2)},\, y_{\min}^{(2)},\, x_{\max}^{(2)},\, y_{\max}^{(2)}]$"
\end{itemize}

<person> = person | subject | individual | human

<persons> = people | subjects | individuals | humans

\end{tcolorbox}

\section{Limitations}
\label{sec:limitations}

% Due to the nature of how VSGaze is constructed, VAT and ChildPlay only contain shared attention annotations towards heads. An important future extension of this benchmark would involve including shared attention annotations towards objects or other items in the scene.

% \agnote{what other limitations? potentially compute limitations? for example for training larger Qwen models. And for evaluating closed source models on the full test set.}

\mypartitle{Task and Representation Scope.}
We evaluate two third-person tasks with 2D gaze-target outputs. We did not include 3D gaze estimation, which predicts gaze direction from the head or face alone and does not require scene understanding, nor semantic gaze target detection (object of regard), egocentric gaze prediction, or gaze-based intent prediction, which belongs more broadly to social behavior understanding. The 2D point is only the output and evaluation representation: predicting it requires reasoning about head orientation and, implicitly, 3D scene structure. Explicit 3D gaze-target evaluation would be valuable but currently lacks suitable datasets. Although Qwen3-VL is trained for coordinate-based grounding and our grid-based control (Appendix~\ref{sec:grid_gaze}) does not improve results, continuous coordinate outputs do not fully isolate gaze inference from coordinate decoding, especially for VLMs without grounding training; region-selection protocols are a complementary direction.

\mypartitle{Frame-Based Evaluation.}
VAT and ChildPlay are video datasets, but we evaluate single frames. This follows the dominant protocol: most methods reported on VAT are frame-based, and temporal models have so far shown no or very small gains. Temporal context may still help in ambiguous, occluded, or dynamic scenes, and evaluating video VLMs is an important extension; we do not expect it alone to close the large zero-shot gap observed here.

\mypartitle{Query Design and Training.}
Each query addresses one task and one subject or pair. Classifying all pairs in a scene therefore requires one query per pair and task, which grows quickly with the number of people; joint multi-task and multi-subject querying would be more efficient. Our fine-tuning uses only answer supervision, without reasoning traces (Appendix~\ref{app:qa_examples}).

\mypartitle{Social Gaze Data.}
Due to the construction of VSGaze, the shared attention annotations in VAT and ChildPlay are limited to attention shared toward human heads. As a result, our benchmark does not yet evaluate shared attention toward objects or other scene entities. Extending the annotation space to include object-level and more general scene-level shared attention would provide a more comprehensive evaluation of social gaze understanding.

\mypartitle{Cost and Optimization.}
VLMs are substantially larger than specialized gaze models, which leads to higher hardware requirements during both fine-tuning and inference, especially in terms of GPU memory. Their autoregressive next-token prediction mechanism also makes inference slower than direct prediction models, and this cost becomes more pronounced when using thinking or reasoning modes, where a complete prediction may take several seconds. A promising direction is therefore to design more efficient architectures that better exploit VLM features while reducing the computational overhead for gaze-specific prediction.

In addition, when fine-tuning VLMs for gaze following and social gaze prediction, the supervision is applied through text-based language modeling losses rather than direct numerical losses on gaze coordinates or social gaze labels. This makes the optimization objective less directly aligned with the target evaluation metrics than in specialized gaze models. Whether language-modeling-based optimization can provide gradients that are sufficiently consistent with task-specific spatial and semantic objectives remains an important question for future research.

Finally, evaluating closed-source VLMs at scale is costly because inference requires paid API calls, and the strongest models are often expensive to query. Therefore, we evaluate closed-source models on 1k subsets for each task. Although this setting provides a practical comparison under realistic budget constraints, larger-scale evaluation would yield more robust and comprehensive estimates of their gaze understanding performance.

\section{Broader Impact}
\label{sec:broader-impact}

Understanding human gaze behavior has several positive societal applications, including the development of automated tools to support clinicians in tasks such as autism screening \cite{lord_autism_1989,lord_autism_2012}, where gaze patterns can provide important diagnostic cues. Because VLMs often offer easy-to-use interfaces, our study can help practitioners better understand the capabilities and limitations of current models when applied to gaze-related tasks.

At the same time, the increasing accessibility of such tools raises important ethical considerations. Gaze data can reveal sensitive information about attention, preferences, health, and cognitive state, creating risks related to privacy, surveillance, and misuse. These tools should therefore only be used in appropriate contexts, and with the prior informed consent of the individuals being observed.

%%%%%%%%%%%%%%%%%%%%%%%%%%%%%%%%%%%%%%%%%%%%%%%%%%%%%%%%%%%%

\newpage
\section*{NeurIPS Paper Checklist}

\begin{enumerate}

\item {\bf Claims}
    \item[] Question: Do the main claims made in the abstract and introduction accurately reflect the paper's contributions and scope?
    \item[] Answer: \answerYes{}{} % Replace by \answerYes{}, \answerNo{}, or \answerNA{}.
    \item[] Justification: The abstract and introduction (Section~\ref{sec:introduction}) state the claims and the scope of the benchmark (two third-person, frame-based gaze tasks), which match the results in Section~\ref{sec:experimentalprotocol} onward.

\item {\bf Limitations}
    \item[] Question: Does the paper discuss the limitations of the work performed by the authors?
    \item[] Answer: \answerYes{} % Replace by \answerYes{}, \answerNo{}, or \answerNA{}.
    \item[] Justification: Limitations are detailed in Appendix~\ref{sec:limitations}.

\item {\bf Theory assumptions and proofs}
    \item[] Question: For each theoretical result, does the paper provide the full set of assumptions and a complete (and correct) proof?
    \item[] Answer: \answerNA{} % Replace by \answerYes{}, \answerNo{}, or \answerNA{}.
    \item[] Justification: We do not have theoretical results in the paper.

    \item {\bf Experimental result reproducibility}
    \item[] Question: Does the paper fully disclose all the information needed to reproduce the main experimental results of the paper to the extent that it affects the main claims and/or conclusions of the paper (regardless of whether the code and data are provided or not)?
    \item[] Answer: \answerYes{} % Replace by \answerYes{}, \answerNo{}, or \answerNA{}.
    \item[] Justification: Please refer to the training setup in Appendix~\ref{sec:appendixadaptation} and the prompts (Appendix~\ref{app:prompts_gf} and \ref{app:prompts_sg}) we used.

\item {\bf Open access to data and code}
    \item[] Question: Does the paper provide open access to the data and code, with sufficient instructions to faithfully reproduce the main experimental results, as described in supplemental material?
    \item[] Answer: \answerYes{} % Replace by \answerYes{}, \answerNo{}, or \answerNA{}.
    \item[] Justification: We will release the data and code upon cleanup.

\item {\bf Experimental setting/details}
    \item[] Question: Does the paper specify all the training and test details (e.g., data splits, hyperparameters, how they were chosen, type of optimizer) necessary to understand the results?
    \item[] Answer: \answerYes{} % Replace by \answerYes{}, \answerNo{}, or \answerNA{}.
    \item[] Justification: Please refer to experimental protocol in Section \ref{sec:experimentalprotocol}.

\item {\bf Experiment statistical significance}
    \item[] Question: Does the paper report error bars suitably and correctly defined or other appropriate information about the statistical significance of the experiments?
    \item[] Answer: \answerNo{} % Replace by \answerYes{}, \answerNo{}, or \answerNA{}.
    \item[] Justification: We do not report error bars because inference is deterministic and each fine-tuning setting is run once due to computational cost.

\item {\bf Experiments compute resources}
    \item[] Question: For each experiment, does the paper provide sufficient information on the computer resources (type of compute workers, memory, time of execution) needed to reproduce the experiments?
    \item[] Answer: \answerYes{} % Replace by \answerYes{}, \answerNo{}, or \answerNA{}.
    \item[] Justification: Please refer to the implement details in Appendix~\ref{sec:appendixadaptation}.

\item {\bf Code of ethics}
    \item[] Question: Does the research conducted in the paper conform, in every respect, with the NeurIPS Code of Ethics \url{https://neurips.cc/public/EthicsGuidelines}?
    \item[] Answer: \answerYes{} % Replace by \answerYes{}, \answerNo{}, or \answerNA{}.
    \item[] Justification:  We have read and observed the NeurIPS Code of Ethics.

\item {\bf Broader impacts}
    \item[] Question: Does the paper discuss both potential positive societal impacts and negative societal impacts of the work performed?
    \item[] Answer: \answerYes{} % Replace by \answerYes{}, \answerNo{}, or \answerNA{}.
    \item[] Justification: Please refer to Section \ref{sec:broader-impact}.

\item {\bf Safeguards}
    \item[] Question: Does the paper describe safeguards that have been put in place for responsible release of data or models that have a high risk for misuse (e.g., pre-trained language models, image generators, or scraped datasets)?
    \item[] Answer: \answerNA{} % Replace by \answerYes{}, \answerNo{}, or \answerNA{}.
    \item[] Justification: The paper poses no such risks.

\item {\bf Licenses for existing assets}
    \item[] Question: Are the creators or original owners of assets (e.g., code, data, models), used in the paper, properly credited and are the license and terms of use explicitly mentioned and properly respected?
    \item[] Answer: \answerYes{} % Replace by \answerYes{}, \answerNo{}, or \answerNA{}.
    \item[] Justification: The original papers for leveraged datasets, code and models have been cited.
    Appropriate licenses will be included when releasing code and data.

\item {\bf New assets}
    \item[] Question: Are new assets introduced in the paper well documented and is the documentation provided alongside the assets?
    \item[] Answer: \answerNA{} % Replace by \answerYes{}, \answerNo{}, or \answerNA{}.
    \item[] Justification: No assets have been currently released with the paper.

\item {\bf Crowdsourcing and research with human subjects}
    \item[] Question: For crowdsourcing experiments and research with human subjects, does the paper include the full text of instructions given to participants and screenshots, if applicable, as well as details about compensation (if any)? 
    \item[] Answer: \answerNA{} % Replace by \answerYes{}, \answerNo{}, or \answerNA{}.
    \item[] Justification: The paper does not involve crowdsourcing nor research with human subjects.

\item {\bf Institutional review board (IRB) approvals or equivalent for research with human subjects}
    \item[] Question: Does the paper describe potential risks incurred by study participants, whether such risks were disclosed to the subjects, and whether Institutional Review Board (IRB) approvals (or an equivalent approval/review based on the requirements of your country or institution) were obtained?
    \item[] Answer: \answerNA{} % Replace by \answerYes{}, \answerNo{}, or \answerNA{}.
    \item[] Justification: The paper does not involve crowdsourcing nor research with human subjects.

\item {\bf Declaration of LLM usage}
    \item[] Question: Does the paper describe the usage of LLMs if it is an important, original, or non-standard component of the core methods in this research? Note that if the LLM is used only for writing, editing, or formatting purposes and does \emph{not} impact the core methodology, scientific rigor, or originality of the research, declaration is not required.
    %this research? 
    \item[] Answer: \answerYes{} % Replace by \answerYes{}, \answerNo{}, or \answerNA{}.
    \item[] Justification: VLMs are central to our study. We evaluate their zero-shot capabilities and fine-tune selected models for gaze following and social gaze prediction. The experimental settings and implementation details are described in the paper.

\end{enumerate}

\end{document}